\documentclass{article} % For LaTeX2e
\usepackage{colm2024_conference}

\usepackage{microtype}
\usepackage{hyperref}
\usepackage{url}
\usepackage{booktabs}

\usepackage{xcolor}         % colors
\usepackage{multirow}
\usepackage{graphicx}
\usepackage{ulem}
\usepackage{caption}
\usepackage{subcaption}
\usepackage{amssymb}
\usepackage{booktabs}
\usepackage{multibib}
\usepackage{colortbl}
\usepackage{makecell}
\usepackage{wrapfig}
\usepackage{color}
\usepackage{float}

\newfloat{figtab}{htb}{fgtb}
\makeatletter
  \newcommand\figcaption{\def\@captype{figure}\caption}
  \newcommand\tabcaption{\def\@captype{table}\caption}
\makeatother

\usepackage{pifont}
\usepackage{enumitem}
\setlist{leftmargin=1em}

\usepackage[listings]{tcolorbox}
\tcbuselibrary{listings,theorems}
\newtcolorbox{mybox}{colback=white!5!white,colframe=black!75!black, left=.05in, right=.05in}

\definecolor{bluex}{rgb}{0.27, 0.42, 0.81}
\definecolor{purplex}{HTML}{9564bf}
\definecolor{red3}{HTML}{C52A20}
\definecolor{red2}{HTML}{B36A6F}
\definecolor{red1}{HTML}{FFb5b5}
\definecolor{purple}{HTML}{B36A6F}
\definecolor{darkyellow}{HTML}{D5BA82}
\definecolor{blue1}{HTML}{508AB2}
\definecolor{blue2}{HTML}{C4E4E3}
\definecolor{green1}{HTML}{A1D0C7}
\definecolor{green2}{HTML}{BFF6BA}
\definecolor{green3}{HTML}{028100}
\definecolor{teal}{HTML}{508AB2}
\definecolor{purple1}{HTML}{8d3a94}

\newtcbtheorem[]{exmp}{Example}%
{colback=purple1!5,colframe=purple1!80,fonttitle=\bfseries, left=.02in, right=.02in,bottom=.02in, top=.02in}{exmp}

\title{Learning From Mistakes Makes LLM Better Reasoner}

\author{Shengnan An\thanks{\, Work done during the internship at Microsoft.}\hspace{0.4mm} $^{\diamondsuit,\clubsuit}$,\, Zexiong Ma$^{*\heartsuit,\clubsuit}$,\, Zeqi Lin\thanks{\, Corresponding authors.}\hspace{0.4mm} $^{\clubsuit}$, Nanning Zheng$^{\dagger\diamondsuit}$,\\\textbf{Jian-Guang Lou}$^{\clubsuit}$,\, \textbf{Weizhu Chen}$^{\clubsuit}$\vspace{1mm}\\
  $^{\diamondsuit}$IAIR, Xi'an Jiaotong University,\,\,
  $^{\clubsuit}$Microsoft Corporation,\, 
  $^{\heartsuit}$Peking University\vspace{1mm}\\
  $^{\diamondsuit}$\texttt{\{an1006634493@stu, nnzheng@mail\}.xjtu.edu.cn}, \\
  $^{\heartsuit}$\texttt{mazexiong@stu.pku.edu.cn}, $^{\clubsuit}$\texttt{\{Zeqi.Lin, jlou, wzchen\}@microsoft.com}
}

\colmfinalcopy % Uncomment for camera-ready version, but NOT for submission.
\begin{document}

\maketitle

\begin{abstract}
% This work introduces \textsc{Le}arning from \textsc{M}ist\textsc{a}kes (\textsc{LeMa}) to enhance the math reasoning capability of large language models (LLMs).
Large language models (LLMs) recently exhibited remarkable reasoning capabilities on solving math problems.
To further improve their reasoning capabilities, this work explores whether LLMs can \textsc{Le}arn from \textsc{M}ist\textsc{a}kes (\textsc{LeMa}), akin to the human learning process.
Consider a human student who failed to solve a math problem, he will learn from what mistake he has made and how to correct it.
Mimicking this error-driven learning process, 
\textsc{LeMa} incorporates mistake-correction data pairs during fine-tuning LLMs.
% \textsc{LeMa} fine-tunes LLMs on mistake-correction data pairs.
Specifically, we first collect inaccurate reasoning paths from various LLMs, and then employ GPT-4 as a ``corrector'' to identify the mistake step, explain the reason for the mistake, correct the mistake and generate the final answer.
In addition, we apply a correction-centric evolution strategy that effectively expands the question set for generating correction data.
Experiments across various LLMs and reasoning tasks show that \textsc{LeMa} effectively improves CoT-alone fine-tuning.
Our further ablations shed light on the non-homogeneous effectiveness between CoT data and correction data.
% Our further analysis sheds light on the non-homogeneous effectiveness between CoT data and correction data, and the contribution from different correction information.
% Across five open-source LLMs and five challenging reasoning tasks, 
% \textsc{LeMa} consistently improves the reasoning performance compared with fine-tuning on CoT data alone.
% Impressively, \textsc{LeMa} can also benefit specialized LLMs such as WizardMath and MetaMath, improving the accuracy from 82.3\% to 85.4\% on GSM8K and from 22.7\% to 27.1\% on MATH.
These results suggest a significant potential for LLMs to improve through learning from their mistakes.
Our code, models and prompts are publicly available at \href{https://github.com/microsoft/LEMA}{Github Link}.
\end{abstract}

\begin{figure*}[h]
    \centering
    \includegraphics[width=.9\textwidth]{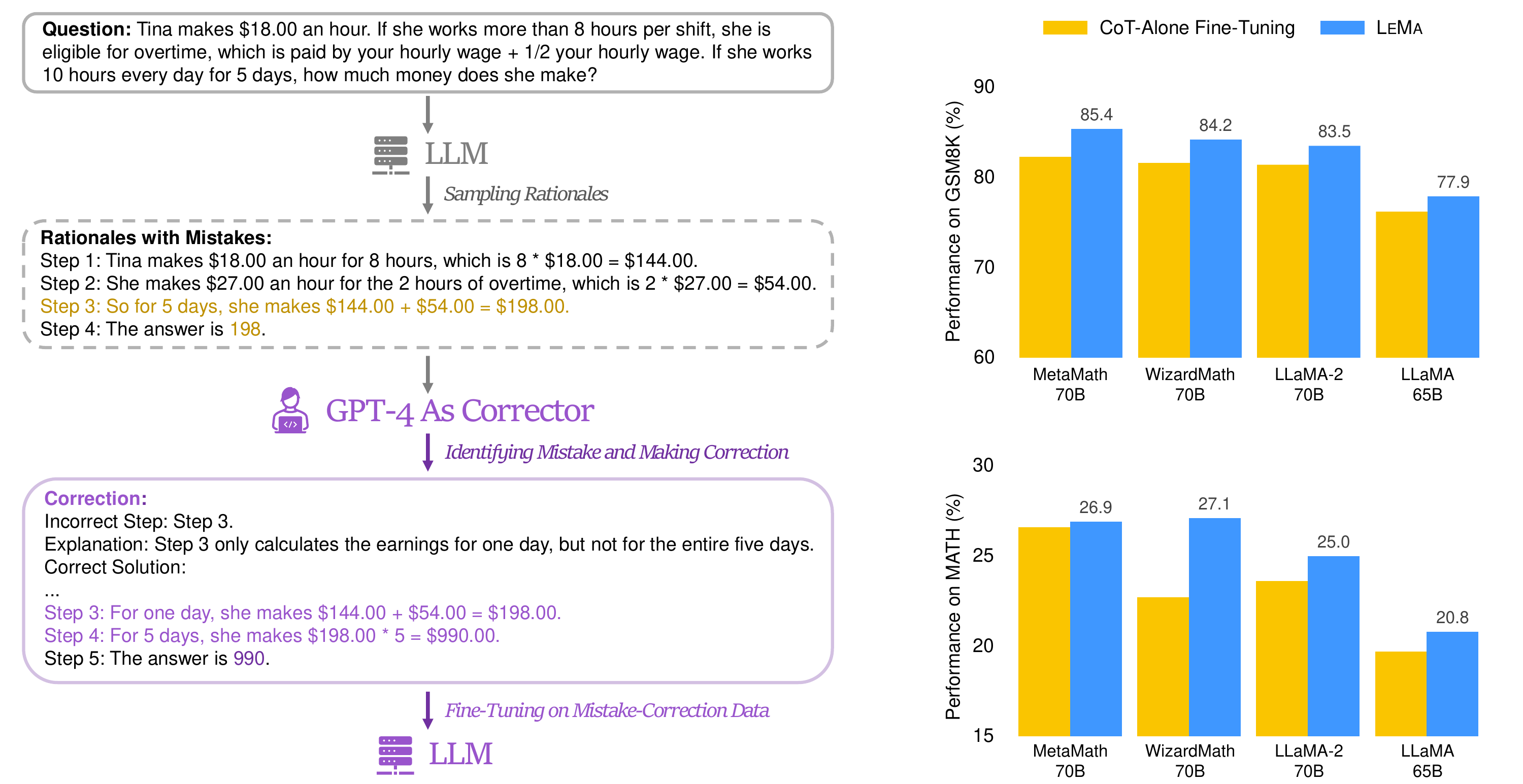}
    \caption{
    Left: Process of \textsc{Le}arning from \textsc{M}ist\textsc{a}kes (\textsc{LeMa}).
    Right: Performance of \textsc{LeMa} on GSM8K and MATH.
    }
    \label{fig:overall}
\end{figure*}

\section{Introduction}

\begin{center}
    \fcolorbox{white}{white}{\parbox{.9\linewidth}{
    \centerline{\textit{Mistakes are the portals of discovery.}}
    \rightline{\textit{---James Joyce}}
}}
\end{center}

With exponential growth in data size and model scale, contemporary large language models~\citep{brown2020language, zhang2022opt, hoffmann2022training, smith2022using, openai2023gpt4, anil2023palm} have demonstrated significant advancements on various NLP tasks, particularly in mathematical problem solving that necessitates complex chain-of-thought (CoT) reasoning~\citep{wei2022chain, wang2022self, li2023making, shi2023language, qin2023chatgpt, lightman2023lets}.
In terms of performance on challenging mathematical tasks like GSM8K~\citep{cobbe2021training} and MATH~\citep{hendrycks2021measuring}, proprietary large language models, including GPT-4~\citep{openai2023gpt4} and PaLM-2~\citep{anil2023palm}, have attained notable results.
However, open-source LLMs such as LLaMA-2~\citep{touvron2023llama2} still have much room for improvement.

To further improve the CoT reasoning capabilities of open-source LLMs for tackling mathematical tasks,
% performance of open-source LLMs on reasoning tasks, 
a common approach is to fine-tune these models using annotated/generated question-rationale data pairs (referred to as \textbf{CoT data}), which directly teach the model how to perform CoT reasoning on these tasks~\citep{magister2022teaching, huang2022large, ho2022large, li2022explanations, yuan2023scaling, luo2023wizardmath, yu2023metamath, li2023query, liang2023let, ranaldi-freitas-2024-aligning}.
While this straightforward learning process has exhibited its effectiveness, this study investigates whether the reasoning capabilities of LLMs can be further improved through a backward learning process, i.e., learning from the mistakes that LLMs have made.
The insight of learning from mistakes comes from the learning process of human students.
Consider a student who is just beginning to learn math.
Beyond learning from golden knowledge and examples in books, he also does exercises.
After failing to solve a problem, he will learn what mistakes he made and how to correct them.
By learning from the mistakes he has made, his reasoning capability will be further improved.
Inspired by this error-driven learning process, this work explores whether the reasoning capabilities of LLMs can also benefit from understanding and correcting mistakes.

To this end, we first generate mistake-correction data pairs (referred to as \textbf{correction data}) and then inject these correction data into the CoT fine-tuning process (Figure~\ref{fig:overall}).
% fine-tune LLMs using correction data .
For generating correction data, we employ multiple LLMs, including the LLaMA and GPT series models, to collect inaccurate reasoning paths (i.e., with incorrect final answers).
We then use GPT-4 as the ``corrector'' to generate corrections for these inaccurate reasoning paths.
The generated corrections contain three pieces of information:
(1) the incorrect step in the original solution,
(2) an explanation of why this step is incorrect,
and (3) how to correct the original solution to arrive at the correct final answer.
After filtering out corrections with incorrect final answers, our human evaluation reveals that our correction data exhibits adequate quality for the subsequent fine-tuning stage.
In addition to using the original training questions to generate correction data, we also consider extending the question sets to scale up our correction data.
Inspired by the evolution techniques for CoT data~\citep{xu2023wizardlm, yu2023metamath, li2023query}, we apply a correction-centric evolution strategy:
compared to randomly selecting seed questions for evolution, our correction-centered evolution focuses more on moderately difficult questions for expanding the correction data.
% which focus on moderate difficulty questions for expanding correction data, rather than randomly choosing questions for evolving.
We blend the generated correction data with the CoT data and then fine-tune LLMs to perform \textsc{Le}arning from \textsc{M}ist\textsc{a}kes (\textsc{LeMa}).
% We fine-tune LLMs on a mixture of CoT data and correction data to perform \textsc{Le}arning from \textsc{M}ist\textsc{a}kes (\textsc{LeMa}).
% We perform \textsc{Le}arning from \textsc{M}ist\textsc{a}kes (\textsc{LeMa}) through fine-tuning LLMs on both CoT data and correction data. 
% QLoRA~\citep{dettmers2023qlora}.

Our experiments on five open-source LLMs and five challenging reasoning tasks demonstrate the effectiveness of \textsc{LeMa}.
Compared to fine-tuning on CoT data alone, \textsc{LeMa} consistently improves the performance across various LLMs and tasks.
For instance, \textsc{LeMa} with LLaMA-2-70B~\citep{touvron2023llama2} achieves 83.5\% on GSM8K and 25.0\% on MATH, while fine-tuning on CoT data alone yields 81.4\% and 23.6\%, respectively.
By incorporating our correction-centric evolution strategy on MATH, \textsc{LeMa} with LLaMA-2-70B can be further improved from 25.0\% to 29.3\%.
Moreover, \textsc{LeMa} can also enhance specialized LLMs such as WizardMath~\citep{luo2023wizardmath} and MetaMath\citep{yu2023metamath}.
% Moreover, \textsc{LeMa} is also compatible with specialized LLMs such as WizardMath~\citep{luo2023wizardmath} and MetaMath\citep{yu2023metamath}.
% Impressively, \textsc{LeMa} improves the performance of MetaMath-70B from 82.3\% to 85.4\% on GSM8K and improves WizardMath-70B from 22.7\% to 27.1\% on MATH.
% For instance, \textsc{LeMa} can improves the performance of MetaMath-70B~\citep{yu2023metamath} 
% \textsc{LeMa} with WizardMath-70B~\citep{luo2023wizardmath}/MetaMath-70B~\citep{yu2023metamath} achieves 84.2\%/85.4\% pass@1 accuracy on GSM8K and 27.1\%/26.9\% on MATH.
% This surpasses the SOTA performance achieved by non-execution\footnote{We mainly compare \textsc{LeMa} with non-execution models which do not rely on the external executor to get final answers.} open-source models on these challenging tasks.
In addition to math tasks, \textsc{LeMa} also benefits commonsense reasoning, improving the performance of LLaMA-2-70B on CSQA~\citep{commonsenseqa2019} from 84.2\% to 85.3\%.

% After the preliminary inspection of the feasibility of \textsc{LeMa}, our analysis on the effectiveness of correction data sheds further light:

Beyond these impressive results, our ablation studies on correction data shed further light.
In controlling the training data sizes and training tokens to be the same, our experimental results reveal that mixing CoT and correction data outperforms a single data source.
These results indicate the non-homogeneous effectiveness of CoT data and correction data.
Moreover, compared with randomly selecting seed questions, our correction-centric evolution better improves the performance of \textsc{LeMa}.
It demonstrates that moderately difficult questions are more suitable for expanding the correction data.

\begin{figure*}[t]
    \centering
    \includegraphics[width=.99\textwidth]{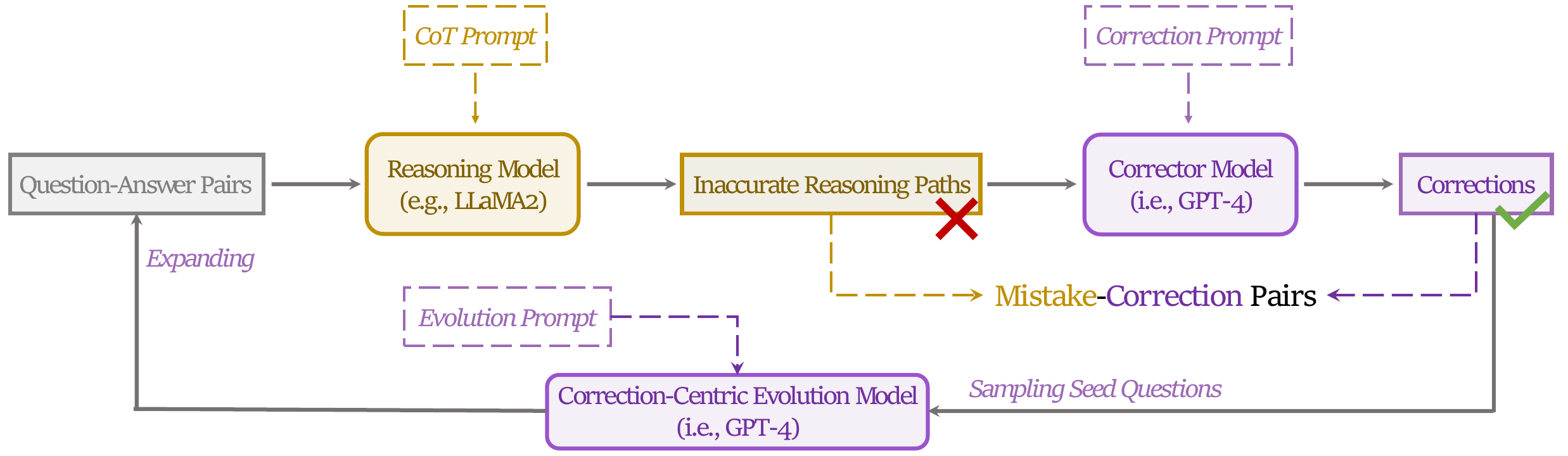}
    \caption{
    Process of generating and expanding correction data.
    }
    \label{fig:corection_pipeline}
\end{figure*}

% Compared with this kind of auxiliary data, our correction data is more informative:
% beyond judging the quality of reasoning steps, our correction data teaches the model to explain its judgment and to revise the original reasoning path according to the judgement.

\section{Methodology}

% \textsc{LeMa} consists of two primary stages: generating correction data and fine-tuning LLMs.
% Figure~\ref{fig:overall} (left) presents an overview of 
\textsc{LeMa} consists of three primary stages: generating correction data, correction-centric evolution, and fine-tuning.

\subsection{Correction Data Generation}\label{sec:method_data}

Figure~\ref{fig:corection_pipeline} briefly illustrates the process of generating correction data.
Given a question-answer example $(q_{i}, {a_{i}}) \in \mathcal{Q}$, a corrector model $\mathcal{M}_c$, and a reasoning model $\mathcal{M}_r$, we generate the mistake-correction data pair $(q_{i}\oplus \widetilde{r_{i}}, c_{i}) \in \mathcal{C}$, where $\widetilde{r_{i}}$ represents an inaccurate reasoning path to the question $q_{i}$, and $c_{i}$ denotes the correction for $\widetilde{r_{i}}$.

\paragraph{Collecting Inaccurate Reasoning Paths.}
We first sample multiple reasoning paths for each question $q_{i}$ using the reasoning model $\mathcal{M}_r$ and retain paths not achieving the correct final answer $a_{i}$,
% We first sample multiple reasoning paths for each question $q_{i}$ using the reasoning model $\mathcal{M}_r$, then just leave paths that do not lead to the correct final answer $a_{i}$,
\begin{equation}\label{sec:sample_mistake}
    \widetilde{r_{i}} \sim \mathcal{M}_r (\mathcal{P}_{r} \oplus q_{i}), \quad \mathrm{Ans}(\widetilde{r_{i}}) \neq a_{i},
\end{equation}
where $\mathcal{P}_{r}$ is the few-shot prompt instructing the model to perform CoT reasoning, and $\mathrm{Ans}(\cdot)$ extracts the final answer from the reasoning path.

\begin{figure*}[t]
    \centering
    \includegraphics[width=.99\textwidth]{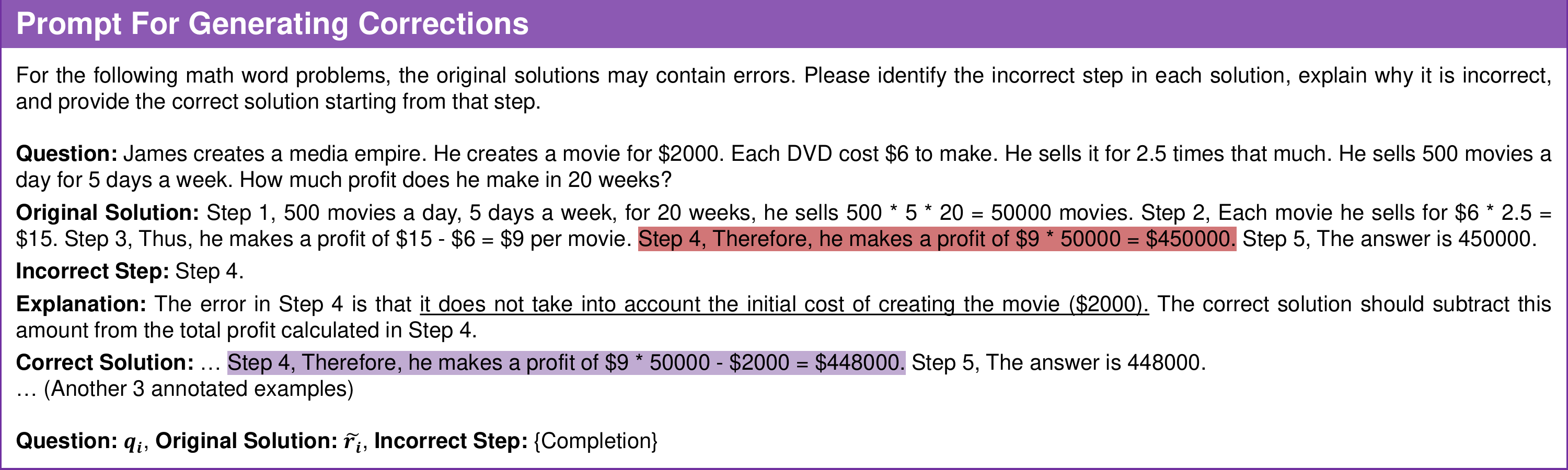}
    \caption{
    A brief illustration of our prompt for generating corrections, containing the \fcolorbox[HTML]{D17677}[HTML]{D17677}{incorrect step} in the original solution, the \underline{reason of mistake}, and the \fcolorbox[HTML]{C0ABD2}[HTML]{C0ABD2}{corrected step}.
    }
    \label{fig:prompt_example}
\end{figure*}

\paragraph{Generating Corrections for Mistakes.}
For question $q_{i}$ and the inaccurate reasoning path $\widetilde{r_{i}}$, we employ the corrector model $\mathcal{M}_c$ to generate a correction and check the final answer in the correction,
\begin{equation}\label{sec:sample_correction}
    c_{i} \sim \mathcal{M}_c (\mathcal{P}_{c} \oplus q_{i} \oplus \widetilde{r_{i}}), \quad \mathrm{Ans}(c_{i}) = a_{i},
\end{equation}
where $\mathcal{P}_{c}$ contains 4 annotated mistake-correction examples to guide the corrector model what kind of information should be contained in the generated corrections.
Figure~\ref{fig:prompt_example} briefly illustrates $\mathcal{P}_{c}$.
Specifically, the annotated corrections comprises three pieces of information:
\begin{itemize}
    \vspace{-2mm}
    \item \textbf{Incorrect Step}: which step in the original reasoning path has made a mistake.
    % \vspace{-1mm}
    \item \textbf{Explanation}: explain what kind of mistake has been made in this step.
    % \vspace{-1mm}
    \item \textbf{Correct Solution}: how to revise the original reasoning path to achieve the correct answer.
\end{itemize}

\paragraph{Human Evaluation for Generated Corrections.}
Before generating data on a large scale, we first manually assess the quality of the generated corrections.
We take LLaMA-2-70B as $\mathcal{M}_r$, utilize GPT-4 as $\mathcal{M}_c$, and generate 50 mistake-correction data pairs based on the GSM8K training set.
We classify the corrections into three quality levels.
\begin{itemize}
    \vspace{-2mm}
    \item \textbf{Excellent}: the corrector successfully identifies the incorrect step in $\widetilde{r_{i}}$, provides a reasonable explanation, and the corrected reasoning path exhibits high continuity with the pre-steps in the original reasoning path\footnote{The high continuity means that the corrected reasoning steps follow the pre-steps generated before the identified mistake step.}.
    % \vspace{-1mm}
    \item \textbf{Good}: the corrector successfully identifies the incorrect step in $\widetilde{r_{i}}$, provides a reasonable explanation, while the corrected reasoning path has minor issues in continuity.
    % \vspace{-1mm}
    \item \textbf{Poor}: the corrector fails to identify the incorrect step in $\widetilde{r_{i}}$ or provides unreasonable explanations.

\end{itemize}
Appendix~\ref{sec:ap_human_evaluation} lists several examples under each quality level.
Our evaluation finds that 35 out of 50 generated corrections are of excellent quality, 11 are good, and 4 are poor.
Based on this human evaluation, we suppose the overall quality of corrections generated with GPT-4 is sufficient for the further fine-tuning stage.
We generate corrections on a large scale and take all corrections that have correct final answers for fine-tuning LLMs.
We provide further analysis on the choice and behavior of corrector model in Section~\ref{sec:ana_corrector}.

\subsection{Correction-Centric Evolution}\label{sec:evolve_method}
After building up the data generation pipeline, we explore how to scale up our correction data.
We consider that expanding the question-answer set $\mathcal{Q}$ is a promising direction, as it primarily determines the correction data diversity.

Inspired by the recent success of evolution techniques on CoT augmentation~\citep{xu2023wizardlm, yu2023metamath, li2023query}, we explore how to effectively apply the evolution method to expand our correction data.
The ``evolution'' means to generate a set of new question-answer pairs from the given \textit{seed questions} by prompting powerful LLMs.

The general evolution method for CoT augmentation randomly selects seed questions to evolve.
However, this strategy does not well suit the nature of our correction data, as too simple or too challenging questions are less valuable for evolving and collecting correction information.
\begin{itemize}
    \vspace{-2mm}
    \item For too simple questions, the reasoning models such as LLaMA can already solve them.
    Evolving these questions may not be effective for collecting mistakes.
    % \vspace{-1mm}
    \item For too challenging questions, the most powerful LLMs still cannot handle them.
    Evolving these questions may lead to much inaccurate information in corrections.
\end{itemize}
Therefore, we apply a \textbf{correction-centric evolution} strategy which more focuses on moderately difficult questions:
\textit{we only sample seed questions that occur in our correction data $\mathcal{C}$,} rather than randomly sampling from the entire set $\mathcal{Q}$,
\begin{equation}
    \hat{q_{i}} \sim \mathcal{M}_e(\mathcal{P}_{e} \oplus q_{i}), \quad q_{i}\in\mathcal{C},
\end{equation}
where $q_{i}$ is the seed question, and $\mathcal{M}_e$ and $\mathcal{P}_{e}$ are the LLM and prompt for evolving questions, respectively.
Appendix~\ref{sec:ap_evolve_prompt} illustrates our $\mathcal{P}_{e}$.

The underlying principle of this strategy is straightforward.
% questions included within the correction data tend to exhibit a moderate level of complexity.
% The idea behind this strategy is quite simple: questions that appear in the correction data are more likely to be of moderate difficulty.
If one question frequently appears in correction data, it means that this question is not well solved by many reasoning models, but its inaccurate reasoning paths can be well handled by the corrector model.
% In addition, if a question appears more frequently in the correction data, it indicates that this question is more valuable for teaching reasoning models.

% If one question appears in correction data, it means that this question is not well solved by all reasoning models, but its inaccurate reasoning path can be corrected by the corrector model, indicating that it has a moderate difficulty.

% For questions that appear in correction data, they are not well solved by all reasoning models bu can be corrected by the corrector model, indicating that they have a moderate difficulty.

% First, for too simple questions in 

% Specifically, it first rewrites the original question into a new one, then generate the answer for the new question.

% Recently, the evolution techniques has been widely used for CoT augmentation.

% Recent work on CoT augmentation also indicated that the augmenting questions has a higher upper-bound than
% there are mainly two directions to scale up our data:
% expanding the question-answer set, or collecting more inaccurate reasoning paths.
% We consider that expanding the question-answer set is the most promising direction, as it contributes to the diversity of the correction data at the source point.
% Recent work on CoT augmentation also indicated that the augmenting questions has a higher upper-bound than 

% As introduced in Section~\ref{sec:method_data}, our correction data is generated based on the given question-answer set .

\subsection{Fine-Tuning LLMs}\label{sec:fine_tuning}
After generating the correction data, we fine-tune LLMs to examine whether these correction data can facilitate CoT reasoning.
% with the help of QLoRA\footnote{ \href{https://github.com/artidoro/qlora}{https://github.com/artidoro/qlora}.}.
We compare the results under two settings:
\begin{itemize}
    \vspace{-2mm}
    \item \textbf{Fine-Tuning on CoT Data Alone}.
    % We fine-tune the model on question-rationale data alone.
    In addition to the annotated data in each task, we additionally take CoT data augmentation following existing methods~\citep{yuan2023scaling, li2023query, yu2023metamath}.
    We generate more reasoning paths for each question in the training sets with GPT-4 and filter out paths with wrong final answers.
    We apply this CoT data augmentation to set up strong fine-tuning baselines that only utilize CoT data.
    % and also facilitates our further ablation study on controlling data size for fine-tuning.
    % (detailed in Section~\ref{sec:results}).
    % \vspace{-2mm}
    \item \textbf{Fine-Tuning on CoT Data + Correction Data}.
    We fine-tune LLMs on both CoT data and generated mistake-correction data.
    This setting is referred to as \textsc{LeMa}.
    % Besides the CoT data, we add our generated mistake-correction data for fine-tuning (this setting is referred to as \textsc{LeMa}).
    \vspace{-2mm}
\end{itemize}
% As the latter setting leads to increments on training data size compared to the former one, we take an ablation study that controls the data size of two settings to be the same (detailed in Section~\ref{sec:results}).
Appendix~\ref{sec:ap_format} shows the input-output formats of CoT data and correction data used for fine-tuning and evaluation.

\section{Experimental Setup}

% In this section, we provides details about our experimental setup.

\begin{table*}[t]
\renewcommand\arraystretch{1.2}
% \Huge
\caption{Our main experimental results (\%) on four mathematical reasoning tasks (GSM8K, MATH, SVAMP and ASDiv) and one commonsense reasoning task (CSQA). Appendix~\ref{sec:ap_best3} and \ref{sec:ap_training_curve} illustrate the performance variances during training.
}
\label{tab:main_results}
\centering
\resizebox{.99\linewidth}{!}{
\begin{tabular}{@{}clccccc@{}}
\toprule
\multirow{2}{*}{Model} & \multicolumn{1}{c}{\multirow{2}{*}{Training}} & \multicolumn{5}{c}{Tasks} \\ \cmidrule(l){3-7} 
 & \multicolumn{1}{c}{} & GSM8K & MATH & \multicolumn{1}{l}{SVAMP} & ASDiv & CSQA \\ \midrule
\multirow{2}{*}{LLaMA-2-70B~\citep{touvron2023llama2}} & CoT Fine-Tuning & 81.4 & 23.6 & 80.3 & 80.7 & 84.2 \\
 & \cellcolor{gray!25} + Learning From   Mistakes & \cellcolor{gray!25}83.5 (+2.1) & \cellcolor{gray!25}25.0 (+1.4) & \cellcolor{gray!25}81.6 (+1.3) & \cellcolor{gray!25}82.2 (+1.5) & \cellcolor{gray!25}85.3 (+1.1) \\ \midrule
\multirow{2}{*}{LLaMA-65B~\citep{touvron2023llama}} & CoT Fine-Tuning & 76.2 & 19.7 & 71.9 & 77.4 & 83.1 \\
 & \cellcolor{gray!25} + Learning From   Mistakes & \cellcolor{gray!25}77.9 (+1.7) & \cellcolor{gray!25}20.8 (+1.1) & \cellcolor{gray!25}72.8 (+0.9) & \cellcolor{gray!25}77.7 (+0.3) & \cellcolor{gray!25}84.0 (+0.9) \\ \midrule
\multirow{2}{*}{CodeLLaMA-34B~\citep{rozière2023codellama}} & CoT Fine-Tuning & 68.8 & 19.1 & 67.4 & 73.9 & 78.1 \\
 & \cellcolor{gray!25} + Learning From   Mistakes & \cellcolor{gray!25}71.7 (+2.9) & \cellcolor{gray!25}20.4 (+1.3) & \cellcolor{gray!25}72.0 (+4.6) & \cellcolor{gray!25}74.4 (+0.5) & \cellcolor{gray!25}80.8 (+2.7) \\ \midrule
\multirow{2}{*}{LLaMA-2-13B~\citep{touvron2023llama2}} & CoT Fine-Tuning & 62.9 & 12.2 & 58.0 & 67.8 & 80.4 \\
 & \cellcolor{gray!25} + Learning From   Mistakes & \cellcolor{gray!25}65.7 (+2.8) & \cellcolor{gray!25}12.6 (+0.4) & \cellcolor{gray!25}62.0 (+4.0) & \cellcolor{gray!25}71.1 (+3.3) & \cellcolor{gray!25}81.9 (+1.5) \\ \midrule
\multirow{2}{*}{LLaMA-2-7B~\citep{touvron2023llama2}} & CoT Fine-Tuning & 52.6 & 8.7 & 53.0 & 63.8 & 76.9 \\
 & \cellcolor{gray!25} + Learning From   Mistakes & \cellcolor{gray!25}54.1 (+1.5) & \cellcolor{gray!25}9.4 (+0.7) & \cellcolor{gray!25}54.1 (+1.1) & \cellcolor{gray!25}65.5 (+1.7) & \cellcolor{gray!25}78.8 (+1.9) \\ \bottomrule
\end{tabular}
}
\end{table*}

\subsection{Tasks}

We undertake experiments on five challenging reasoning tasks, including four mathematical reasoning tasks (GSM8K, MATH, SVAMP and ASDiv) and one commonsense reasoning task (CSQA)\footnote{Appendix~\ref{sec:ap_data} contains basic statics about the tasks and data.}.
For GSM8K, MATH and CSQA, we generate correction data based on their training sets.
For SVAMP and ASDiv, we take the same training data for GSM8K.

% \vspace{1mm}
\noindent \textbf{GSM8K}~\citep{cobbe2021training} contains high quality linguistically diverse grade school math word problems.
It has 7,473 training examples with CoT and 1,319 test cases.
% We take its training set to augment CoT data and generate correction data and report the evaluation results on test cases.

% \vspace{1mm}
\noindent \textbf{MATH}~\citep{hendrycks2021measuring} examines math reasoning on solving challenging competition mathematics problems.
It contains 7,500 training CoT data and 5,000 test cases.
% We take its training set to augment CoT data and generate correction data and report the evaluation results on test cases.

% \vspace{1mm}
\noindent \textbf{SVAMP}~\citep{svamp2021} consists of questions with short NL narratives as state descriptions.
For evaluation on SVAMP, we use the same training data as for GSM8K and take all 1,000 examples in SVAMP as test cases.

% \vspace{1mm}
\noindent \textbf{ASDiv}~\citep{asdiv2020} is a diverse math dataset in terms of both language patterns and problem types for evaluating.
% It contains 2,305 English math word problems.
For evaluation on ASDiv, we use the same training data as for GSM8K and test on 2,084 examples in ASDiv\footnote{The original ASDiv contains 2,305 examples and we filter out non-numerical examples, detailed in Appendix~\ref{sec:ap_asdiv}.}.
% Additionally, as the training data for GSM8K only contains questions resulting in numerical answers, we filter out questions in ASDiv that have non-numerical answers (detailed in Appendix~\ref{sec:ap_asdiv}).
% Finally, there are 2,084 test cases for our evaluation on ASDiv. 

% \vspace{1mm}
\noindent \textbf{CSQA}~\citep{commonsenseqa2019} is a question answering dataset for commonsense reasoning.
It has 9,741 examples in the training set and 1,221 examples in the dev set.
% Each example contains a question, five candidate answers and one ground-truth answer.
As it does not contain any CoT annotation, we first annotate 4 CoT examples (detailed in Appendix~\ref{sec:ap_csqa}), then take its training set to augment CoT data and generate correction data.

\begin{figure*}[t]
    \centering
    \includegraphics[width=.99\textwidth]{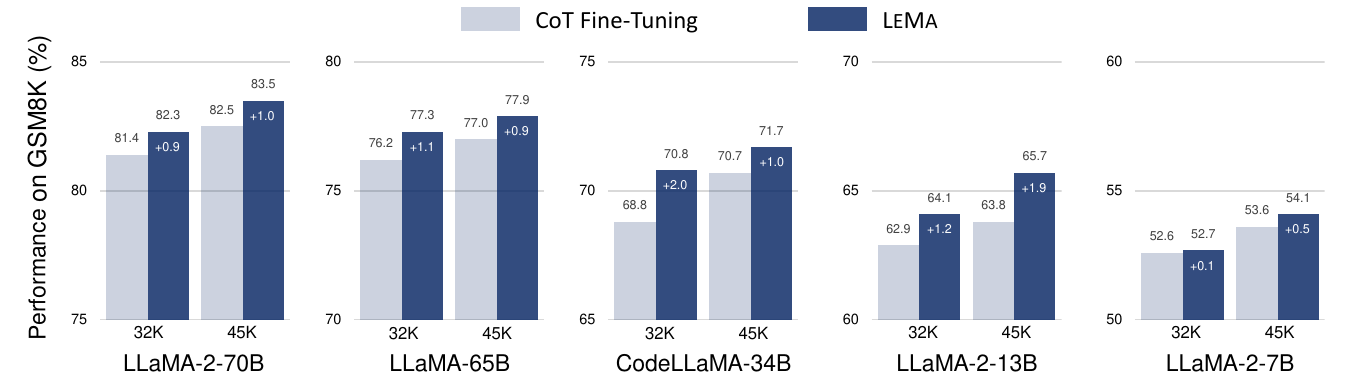}
    \caption{
    Performances of \textsc{LeMa} and CoT-alone fine-tuning with controlled data sizes (32K and 45K) on GSM8K.
    See Table~\ref{tab:training_token} for results with controlled number of training tokens.
    % For each model and each data size, we mark the gain of \textsc{LeMa} compared to CoT-alone fine-tuning.
    }
    \label{fig:datasize}
\end{figure*}

\subsection{Data Construction}

\paragraph{CoT Data.}
For GSM8K (also SVAMP and ASDiv),
the CoT data contains all training examples of GSM8K and 24,948 augmented reasoning paths.
We first generate 30,000 reasoning paths with GPT-4 and filter out 5,052 paths with wrong final answers or unexpected format\footnote{The unexpected format means that the final answer is failed to be extracted from the path with the regular expression.}.
For MATH, the CoT data contains all training examples and 12,509 augmented reasoning paths.
We sample 30,000 reasoning paths with GPT-4 and filter out 17,491 paths.
% In total, there are 20,009 examples for CoT fine-tuning on MATH.
% For CSQA, it does not contain CoT data in original training set, and 
For CSQA, we generate 15,000 reasoning paths with GPT-4 and then filter out 4,464 paths.
% For CSQA, we generate 10,536 reasoning paths for its training examples through few-shot prompting.
% Specifically, we generate 15,000 reasoning paths with GPT-4 and then filter out 4,464 paths.

\paragraph{Correction Data.}
We utilize multiple LLMs to collect inaccurate reasoning paths, including LLaMA-2~\citep{touvron2023llama2}, WizardLM~\citep{xu2023wizardlm}, WizardMath~\citep{luo2023wizardmath}, Text-Davinci-003~\citep{text003}, GPT-3.5-Turbo~\citep{gpt35turbo} and GPT-4~\citep{openai2023gpt4}.
We take GPT-4 as the corrector model.
Finally, we collect 12,523, 6,306, 7,241 mistake-correction pairs based on the training sets of GSM8K, MATH and CSQA, respectively.
% Along with CoT data, there are totally 44,944, 26,315 and 17,777 training examples for \textsc{LeMa} on GSM8K (also SVAMP and ASDiv), MATH and CSQA, respectively.

\paragraph{Correction-Centric Evolution.}
We take 10K bootstrap samples from the questions in our correction data.
We utilize GPT-4 to evolve the questions.
To generate ``ground-truth'' answers for the evolved questions, we utilize GPT-4 to sample three answers for each question and conduct a majority voting.
The question that leads to three different answers will be filtered.
Note that the evolved data will only be used in Section~\ref{sec:exp_evolve}.

\subsection{Fine-Tuning and Evaluation}\label{sec:ft_setting}

We fine-tune multiple open-source LLMs in the LLaMA~\citep{touvron2023llama}, LLaMA-2~\citep{touvron2023llama2}, CodeLLaMA~\citep{rozière2023codellama}, WizardMath~\citep{luo2023wizardmath} and MetaMath~\citep{yu2023metamath} families.
We utilize QLoRA\footnote{\href{https://github.com/artidoro/qlora}{https://github.com/artidoro/qlora}.}~\citep{hu2022lora, dettmers2023qlora} by default to conduct parameter-efficient fine-tuning (PEFT) for these models.
We set low-rank dimension as 64 and dropout rate as 0.05.
We set learning rate as 0.0001 for LLMs larger than (or equal to) 34B and 0.0002 for LLMs smaller than 34B.
% Both max source length and max target length are set as 768.
We set batch size as 96, train for 2,000 steps, and save checkpoints for every 100 training steps.

For evaluation, we evaluate the performance of all saved checkpoints based on vLLM library\footnote{\href{https://github.com/vllm-project/vllm}{https://github.com/vllm-project/vllm}.}~\citep{kwon2023efficient}
and report the accuracy of the best checkpoint.
During inference, we set temperature as 0 (i.e., greedy decoding) and max sample length as 2,048.
To clarify the influence from random disturbances during training, we provide the performances of the best three checkpoints in Appendix~\ref{sec:ap_best3} and the performance curves during the whole training processes in Appendix~\ref{sec:ap_training_curve}.
We do not add demonstration examples into the prompt for both fine-tuning and evaluation by default.
All evaluations are conducted under the same CoT instruction.
For models trained with \textsc{LeMa}, we do not generate corrections during evaluations.
All our experiments can be conducted on 4 x A100 GPU stations.

\begin{table}[t]
\centering
\resizebox{.99\linewidth}{!}{
\parbox{.53\linewidth}{
\centering
\begin{tabular}{@{}ccc@{}}
\toprule
Model & Data & Acc (\%) \\ \midrule
\multirow{2}{*}{LLaMA-2-70B} & CoT-5.8M & 82.1 \\
 & \textsc{LeMa}-5.8M & \textbf{83.5 (+1.4)} \\ \midrule
\multirow{2}{*}{LLaMA-2-13B} & CoT-5.8M & 64.2 \\
 & \textsc{LeMa}-5.8M & \textbf{65.7 (+1.5)} \\ \bottomrule
\end{tabular}
\caption{Performances with the same size of training tokens (5.8M) on GSM8K.}
\label{tab:training_token}
}
\quad
\parbox{.53\linewidth}{
\centering
\begin{tabular}{@{}lc@{}}
\toprule
\multicolumn{1}{c}{Model} & Acc (\%) \\ \midrule
WizardMath-70B~\citep{luo2023wizardmath} & 81.6 \\
WizardMath-70B + \textsc{LeMa} & \textbf{84.2 (+2.6)} \\ \midrule
MetaMath-70B~\citep{yu2023metamath} & 82.3 \\
MetaMath-70B + \textsc{LeMa} & \textbf{85.4 (+3.1)} \\ \bottomrule
\end{tabular}
\caption{Performances of \textsc{LeMa} with specialized LLMs on GSM8K.}
\label{tab:specialized_llm}
}
}
\end{table}

\section{Results and Analysis}\label{sec:results}

We focus on two main research questions in this section.
More results and analysis are contained in Appendix~\ref{ap:more_results}.

\subsection{Can LLMs Learn From Mistakes?}\label{sec:main_results}

% Table~\ref{tab:main_results} and \ref{tab:lead} provide strong experimental evidence that LLMs can learn from mistakes.

% \paragraph{Correction data consistently improves the performance of various LLMs.}
\paragraph{\textsc{LeMa} effectively improves CoT-alone fine-tuning.}
Table~\ref{tab:main_results} shows the main experimental results on five challenging reasoning tasks.
Compared to fine-tuning on CoT data alone, incorporating correction data during fine-tuning brings improvements across all five backbone LLMs and five tasks.
It demonstrates that \textsc{LeMa} can effectively facilicate CoT fine-tuning.
Note that SVAMP and ASDiv can be regarded as two out-of-distribution tasks as the training data is constructed based on GSM8K.
The gains on these two tasks reflect that \textsc{LeMa} has a certain extent of generalizability in the out-of-distribution scenarios.
% Moreover, we notice that the performance of all three checkpoints with correction data can consistently outperforms the best performance with CoT data alone.
% Additionally, Figure~\ref{fig:curve} illustrates the performance curves during the whole training process.
% It shows that our correction data leads to clear improvements during training.
% These consistent improvements demonstrate that the effectiveness of our correction data is robust to the random disturbances during training.

% \paragraph{A more powerful backbone model can learn from mistakes more effectively.}

% For LLaMA-2-70B in Table~\ref{tab:main_results}, it has the highest baseline results while keeping significant improvements (>1\% accuracy gain) across all five tasks.
% For other less powerful models in Table~\ref{tab:main_results}, the gain could sometimes be less significant.
% Such a comparison indicates that the capability of backbone models contributes to how well they can learn from mistakes.

\paragraph{The effectiveness of CoT data and correction data are non-homogeneous.}
If the effectiveness of the two data sources are homogeneous, the gains in Table~\ref{tab:main_results} will be diminished if the data sizes of two fine-tuning settings are controlled as the same.
To further validate the effectiveness of correction data, 
% To validate this,
we conduct two ablation studies with \textbf{controlled data sizes}.
% \textsc{LeMa} and CoT-alone fine-tuning on GSM8K to be the same.
In default settings, we have about 32K examples for CoT-alone fine-tuning and 45K examples for \textsc{LeMa}.
Here are another two controlled settings:
\begin{itemize}
    \vspace{-2mm}
    \item \textsc{LeMa}-32K.
    We keep the 13K correction data and randomly remove 13K CoT data.
    % We reduce the data size of \textsc{LeMa} to be the same with our default CoT-alone fine-tuning.
    % We randomly remove 12,523 examples from the augmented reasoning paths in our CoT data, leaving 32,421 examples for \textsc{LeMa} (including 19,898 CoT data and 12,523 correction data).
    % \vspace{-2mm}
    \item CoT-45K.
    To expand CoT data, we extract the corrected CoT from each correction example.
    % This setting increase the data size of CoT-alone fine-tuning to be the same with our default \textsc{LeMa} setting.
    % For our generated 12,523 correction data, we extract the corrected CoT from the correction information to expand the CoT data.
\end{itemize}
% \vspace{-2mm}
Figure~\ref{fig:datasize} shows that \textsc{LeMa} can still bring gains for four out of five backbone LLMs under the same data size.
It means that these LLMs do learn extra information from our correction data that is not provided by the CoT data.
The only exception is for LLaMA-2-7B.
It indicates that a stronger backbone model can more effectively learn from mistakes.

% \paragraph{Token-level efficiency.}
Despite controlling the training data sizes to be the same, we also investigate the \textbf{training-token efficiency} of \textsc{LeMa} compared with CoT-alone fine-tuning.
Notice that the target-side length of correction data is generally longer than CoT data, so LEMA will have slightly more training tokens than CoT-alone fine-tuning under the same data size.
Specifically, CoT-45K has 5.4M training tokens and \textsc{LeMa}-45K has 5.8M (a $\sim$7\% relative increment).
To conduct the comparison under the same size of training tokens, we construct CoT-5.8M by sampling more reasoning paths (following Section~\ref{sec:fine_tuning}) to add into CoT-45K.

Table~\ref{tab:training_token} shows that \textsc{LeMa} still outperforms CoT-alone fine-tuning with the same number of training tokens.
Note that this comparison is under an unfavorable setup for \textsc{LeMa} as it increases the training samples for CoT-alone fine-tuning.
The improvements in Table~\ref{tab:training_token} further support the non-homogeneous effectiveness of CoT data and correction data.
Moreover, we notice that augmenting more reasoning paths for LLaMA-2-70B does not continuously boost the model performance on GSM8K.
To validate this, we further expand CoT-5.8M to CoT-6.8M and have a 82.2\% accuracy.
Such an observation is in line with the \citet{yu2023metamath}.
We suppose that this is because sampling too many reasoning paths for the same question will only bring redundant information to the training.

% which leads to a slight increment on training tokens under the same data size.

% It is noteworthy that \textsc{LeMa}-32K achieves performances comparable to that of CoT-45K despite having only $\sim81\%$ of the training toknes of CoT-45K.
% It indicates that \textsc{LeMa} also improves the token efficiency compared with using CoT data alone.

% Despite controlling the training data sizes to be the same, 
% It is noteworthy that \textsc{LeMa}-32K achieves performances comparable to that of CoT-45K despite having only $\sim81\%$ of the training toknes of CoT-45K.
% It indicates that \textsc{LeMa} also improves the token efficiency compared with using CoT data alone.

% For LLaMA-2-7B, the performance of two settings are close under the controlled same data size.
% , suggesting that larger models are better at learning from mistakes than smaller ones.
% For LLaMA-2-7B, the performance of \textsc{LeMa} under controlled data size is close to that of CoT-alone fine-tuning.

% It seems that larger language models are better at learning from mistakes than smaller ones.
% Appendix~\ref{ap:more_results} provides further analysis on this.

\paragraph{A stronger backbone model can be more effective at learning from mistakes.}
As evidenced in Table~\ref{tab:main_results}, LLaMA-2-70B has the highest baseline performances in CoT alone fine-tuning, while maintaining significant improvements in all five tasks (an accuracy gain of over 1\%) with the help of \textsc{LeMa}.
In contrast, for other four less powerful models in Table~\ref{tab:main_results}, the improvements from \textsc{LeMa} are occasionally less significant.
This comparison, along with the performance of LLaMA-2-7B in Figure~\ref{fig:datasize}, suggests that the inherent strength of backbone LLMs can influence how well the models can learn from mistakes.

\paragraph{\textsc{LeMa} can also facilitate specialized LLMs.}
To adapt generally pre-trained LLMs into the math domain, there have been several specialized LLMs such as WizardMath~\citep{luo2023wizardmath} and MetaMath~\citep{yu2023metamath}.
We also apply \textsc{LeMa} on these specialized LLMs to further examine its effectiveness.
As these models have been already trained on a large amount of CoT data designed for math tasks, we directly compare \textsc{LeMa} with the results reported in the original papers for these specialized models.
% and report the performance in Table~\ref{tab:lead}.
Table~\ref{tab:specialized_llm} shows that \textsc{LeMa} can further improve these specialized LLMs.
Appendix~\ref{sec:ap_sota} contains detailed comparisons.
% Impressively, \textsc{LeMa} improves the performance of MetaMath-70B from 82.3\% to 85.4\% on GSM8K and improves WizardMath-70B from 22.7\% to 27.1\% on MATH.
% Note that these specialized LLMs are mainly trained on a large scale of CoT data.
% The improvements with specialized LLMs further demonstrate that the effectiveness of CoT data and correction data are non-homogeneous.

% that incorporating correction data can also contribute to the data efficiency.
% We provide further analysis on data efficiency in Section~\ref{sec:further_analysis}.

\begin{figure}[t]
     \centering
     \begin{subfigure}[t]{0.49\textwidth}
        \centering
        \includegraphics[width=.99\textwidth]{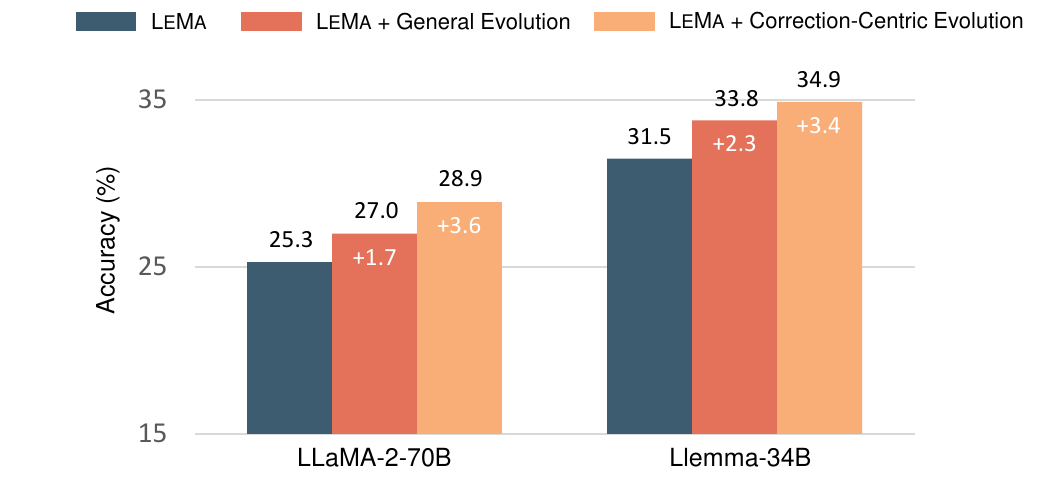}
        \caption{}
        \label{fig:evolve_compare}
     \end{subfigure}
     \hfill
     \begin{subfigure}[t]{0.49\textwidth}
            \centering
            \includegraphics[width=.99\textwidth]{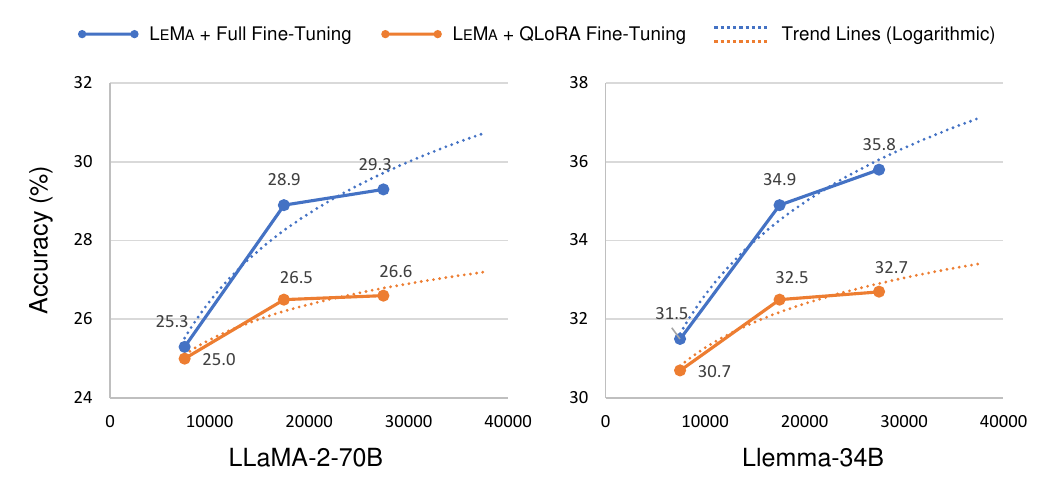}
            \caption{}
            \label{fig:evolve_trend}
     \end{subfigure}
     \caption{Performance of \textsc{LeMa} on MATH with evolution strategies. (a) Compare general and correction-centric evolution strategies (full fine-tuning). (b) The performance trend of \textsc{LeMa} with QLoRA or full fine-tuning. X-axis is the number of sampled questions.
     }
\end{figure}

\subsection{How Beneficial Is Correction-Centric Evolution?}\label{sec:exp_evolve}

Figure~\ref{fig:evolve_compare} and Figure~\ref{fig:evolve_trend} demonstrate further improvements on the performance of \textsc{LeMa} with incorporating the correction-centric evolution strategy to expand the correction data.

\paragraph{Correction-centric evolution can more effectively improve \textsc{LeMa}.}
Figure~\ref{fig:evolve_compare} shows the performance of \textsc{LeMa} with incorporating different evolution strategies.
Besides the correction-centric evolution introduced in Section~\ref{sec:evolve_method}, we also compare with the general evolution strategy applied in previous work~\citep{xu2023wizardlm, yu2023metamath, li2023query}.
For a fair comparison, the number of seed questions is kept the same for both evolution strategies (i.e., 10K).
We also tried the Llemma~\citep{azerbayev2023llemma} model which has been pre-trained on a math-related corpus (such as arXiv papers).
We fully fine-tune LLMs as the correction data scale has been much increased\footnote{Appendix~\ref{sec:ap_ft_set} contains the settings for full fine-tuning.}.

% Table~\ref{fig:evolve_compare} shows the results of \textsc{LeMa} with different evolution strategies.
There are two primary conclusions.
First, \textsc{LeMa} can effectively benefit from evolution techniques.
It indicates that the performance of \textsc{LeMa} can be further improved by incorporating existing data augmentation techniques.
Second, the correction-centric evolution outperforms the general evolution.
It demonstrates that moderately difficult questions are more suitable for expanding the correction data.

\paragraph{Evolution techniques can better facilitate \textsc{LeMa} under full fine-tuning.}
To explore the scaling trend of \textsc{LeMa}, we apply the correction-centric evolution on another 10K sampled seed questions (detailed in Appendix~\ref{sec:ap_another_evolve}).
Figure~\ref{fig:evolve_trend} shows the performance trends of \textsc{LeMa} as the question set expands.
It shows that if only the original question-answer pairs in MATH are used (i.e., the initial points in each line), there is no significant difference in the performances of \textsc{LeMa} between QLoRA and full fine-tuning.
However, as the question set expands, the performance with full fine-tuning improves significantly, while QLoRA fine-tuning increases only slightly.
It indicates that the parameter-efficient fine-tuning can only ``digest'' a limited scale of correction data.
Appendix~\ref{sec:ap_more_analysis} provides further analysis.

\section{Related Work}

\paragraph{LLMs with CoT reasoning.}
\citet{wei2022chain} uncovered the emergence of CoT reasoning capability for extremely large language models, and this reasoning capability was then examined in various reasoning-related domains including logical reasoning~\citep{creswell2022selection, pan2023logic, lei2023boosting}, commonsense reasoning~\citep{commonsenseqa2019, strategyqa2021, ahn2022i}, and math reasoning~\citep{asdiv2020, mawps2016, svamp2021, cobbe2021training, hendrycks2021measuring}.
The impressive performance of LLMs in these domains has spurred the research community to further investigate methods for effectively harnessing and enhancing CoT reasoning for LLMs~\citep{wang2022self, zhou2022least, creswell2022faithful, li2023making, lightman2023lets}.

\vspace{-1mm}
\paragraph{Enhancing CoT reasoning for solving mathematical problems.}
There has been much work dedicated to enhancing the performance of LLMs in solving mathematical problems from various perspectives.
Some studies explored the voting or verification methods based on sampling multiple reasoning paths~\citep{wang2022self, li2023making, lightman2023lets}.
Some methods considered to generate executable programs to obtain the final answer or to integrate plug-in tools that facilitate the execution of external APIs during intermediate steps~\citep{jie2023leveraging, wang2023mathcoder, yue2023mammoth, azerbayev2023llemma, gou2023tora}.
Some work collected math-related corpus such as arXiv papers for pre-training better base models for math~\citep{azerbayev2023llemma, wang2023generative}.
Some work focused on augmenting existing datasets, which expanded training sets or provided external annotations~\citep{magister2022teaching, huang2022large, ho2022large, li2022explanations, luo2023wizardmath, yu2023metamath, li2023query, liang2023let, liu2023tinygsm}.
From the perspective of the techniques used, this work follows the data augmentation approach.

\vspace{-1mm}
\paragraph{Data augmentation for mathematical tasks.}
With the help of advanced LLMs (e.g., GPT-4 and GPT-3.5-Turbo), various methods have been proposed to generate more CoT data for mathematical tasks:
\citet{yuan2023scaling} proposed rejection sampling for augmenting CoT data;
\citet{xu2023wizardlm} evolved the math questions in the training sets;
\citet{li2023query} applied both query augmentation and response augmentation;
\citet{yu2023metamath} used self-verification and FOBAR to generate CoT with high diversity.
While the effectiveness of CoT data has been well studied, how to improve math reasoning with other auxiliary data is still under-explored.
% This work takes a preliminary step toward leveraging auxiliary data:
% we propose to utilize mistake-correction data and experimentally examine its effectiveness.
To this end, there are some preliminary explorations:
\citet{azerbayev2023llemma} and \citet{yue2023mammoth} found that code data can facilitate math reasoning;
\citet{liu2023improving} and \citet{wang2023democratizing} constructed re-ranking data or verification data to make the model judge the quality of reasoning paths.
This work takes a further step toward leveraging auxiliary data:
we propose and examine the effectiveness of mistake-correction data, which informs the model what kind of mistakes could be made in CoT reasoning and how to correct them.

\section{Conclusion}
This work explores whether the reasoning capabilities of LLMs can be further improved by learning from mistakes.
Experimental results and in-depth analysis demonstrate the effectiveness and potential of learning from mistakes.

% % This work explores how to 
% % This work introduces \textsc{LeMa} to help LLMs learn from mistakes.
% % This work explores the potential of LLMs to learn from mistakes.
% This work introduces \textsc{LeMa} that incorporates correction data during fine-tuning.
% Experimental results and in-depth analysis demonstrate the effectiveness and potential of learning from mistakes.
% % By generating correction data and fine-tuning LLMs, our experimental results demonstrate that learning from mistakes can improve the performance of LLMs on challenging reasoning tasks.
% % Our ablation study shed further light on the effectiveness and potential of \textsc{LeMa}.

\section*{Ethics Statement}
Due to the utilization of pre-trained language models, this work could be exposed to some potential risks of ethical issues on general deep learning models (such as social bias and privacy breaches).
We hope that the idea of learning from mistakes would facilitate the development of responsible AI models, for instance, on training LLMs to recognize and modify risky generated contents.
% As explored in this work that the model behavior can be hugely influenced by the provided context, we call for further investigation into how ethical issues can be avoided by controlling the provided context.

\section*{Reproducibility Statement}
We open source our training code, evaluation scripts and fine-tuned checkpoints to facilitate further explorations on learning from mistakes.
For generating the training data, we provide all our prompts used for data generation.
% , and will make our generated data publicly available after the internal data review process..

\section*{Acknowledgments}
Shengnan An and Nanning Zheng were supported in part by NSFC under grant No. 62088102.
Thank Chen Li at IAIR, Xi'an Jiaotong University for his valuable comments on this work.

\bibliography{colm2024_conference}
\bibliographystyle{colm2024_conference}

\newpage
\appendix

This is the Appendix of the paper: \textit{Learning From Mistakes Makes LLM Better Reasoner}.

\section{Discussion}

Here, we discuss further about the insights from our exploration on learning from mistakes.

\subsection{LLMs for Self-Correction}

Recently, much work has investigated the behavior of advanced LLMs (e.g., GPT-4) on correcting mistakes generated by themselves~\citep{valmeekam2023large, stechly2023gpt4, huang2023large}.
We also conduct further analysis on self-correction performance based on our correction data (detailed in Appendix~\ref{sec:ana_corrector}).
These work and our analysis drew the same conclusion: the most powerful LLMs by now still struggle to perform self-correction.
To achieve more reliable utilization of self-correction, we think that there are mainly three directions.
(1) Inject external supervision to verify the correcting process, such as using the labeled final answers (which is applied in our work) or incorporating human feedback.
(2) Train a process-based verifier to judge the quality of self-correction process. \citet{lightman2023lets} has demonstrated the great potential of verifier-based method.
(3) Develop trust-worth LLMs that can at least honestly tell us what it can solve and what does not.

\subsection{Training with Feedback}

To align the behavior of LLMs with human expectations, existing work has tried to collect feedback for the model-generated contents and inject these feedback back into the model through various techniques, such as PPO~\citep{lu2022quark}, RLHF~\citep{openai2023gpt4} and DPO~\citep{rafailov2023direct}.
To reduce human efforts on annotation, some recent work tried to use LLMs to generate feedback, such as RLAIF~\citep{lee2023rlaif}.
From this view, \textsc{LeMa} can also be regarded as injecting the feedback from more powerful LLMs (i.e., GPT-4) into smaller models (e.g., LLaMA).
We highlight one difference here: the injection process of \textsc{LeMa} is just implemented with instruction-based fine-tuning rather than RL-based methods.
It sheds light that for large pre-trained models, it can directly and effectively learn from the comparison between unexpected and expected contents through the input-output fine-tuning process.
This can much save the researchers effort to specially design the learning algorithms.

\subsection{Learning From The World Model}

Recent advancements in LLMs have enabled them to perform a step-by-step approach in problem-solving.
However, this multi-step generation process does not inherently imply that LLMs possess strong reasoning capabilities, as they may merely emulate the superficial behavior of human reasoning without genuinely comprehending the underlying logic and rules necessary for precise reasoning.
This incomprehension results in mistakes during the reasoning process and necessitates the assistance of a ``world model'' that possesses a consciousness prior about the logic and rules governing the real world.
From this perspective, our \textsc{LeMa} framework employs GPT-4 as a ``world model'' to teach smaller models in adhering to these logic and rules, rather than merely mimicking the step-by-step behavior.

\section{Additional Examples}

\subsection{Examples in Human Evaluation}\label{sec:ap_human_evaluation}

\begin{figure*}[t]
    \centering
    \includegraphics[width=.99\textwidth]{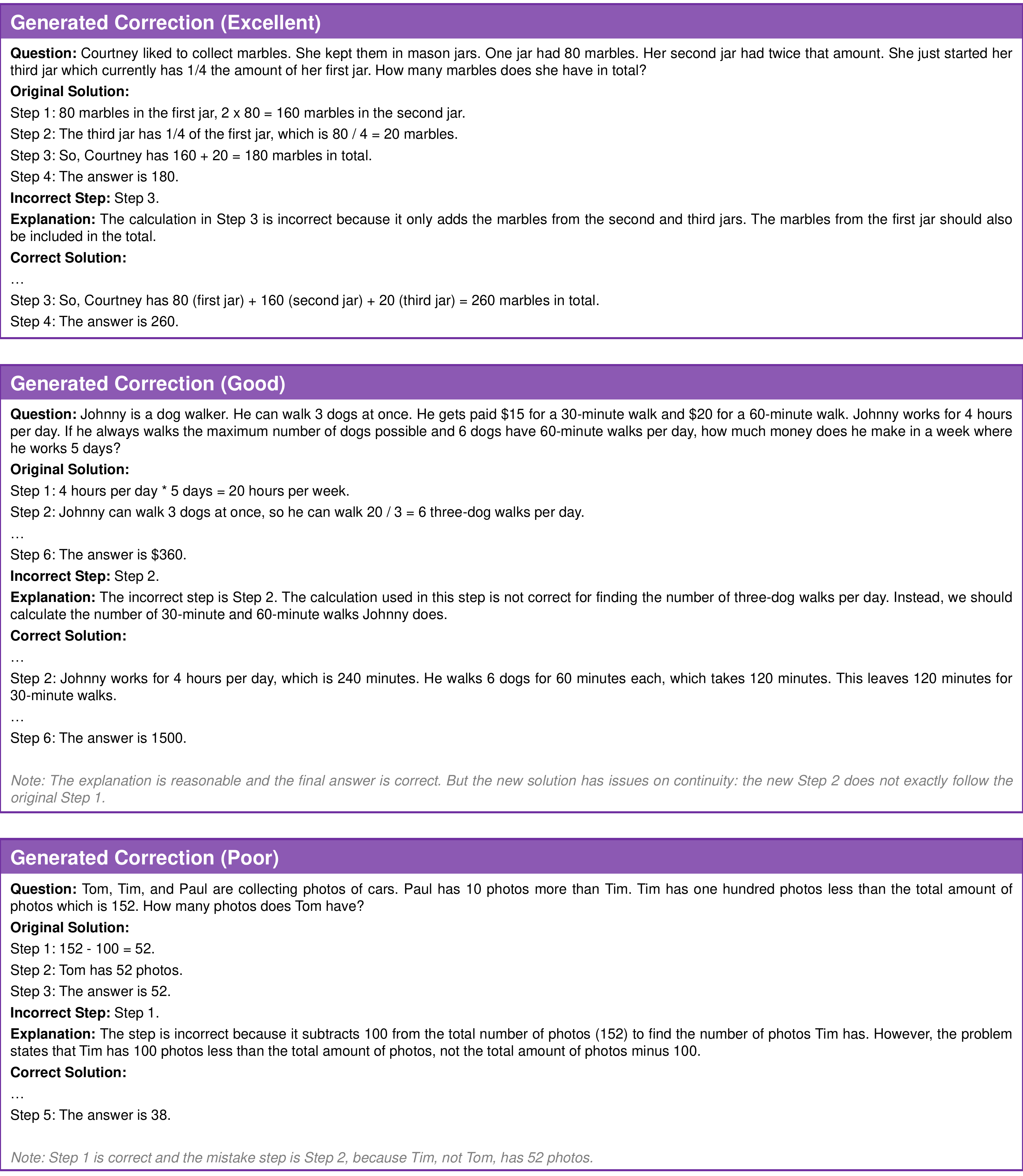}
    \caption{
    Some examples of generated corrections and their quality levels under our human evaluation.
    }
    \label{fig:human_evaluation_examples}
\end{figure*}

Figure~\ref{fig:human_evaluation_examples} illustrates the quality levels of three example corrections.

\begin{figure*}[t]
    \centering
    \includegraphics[width=.99\textwidth]{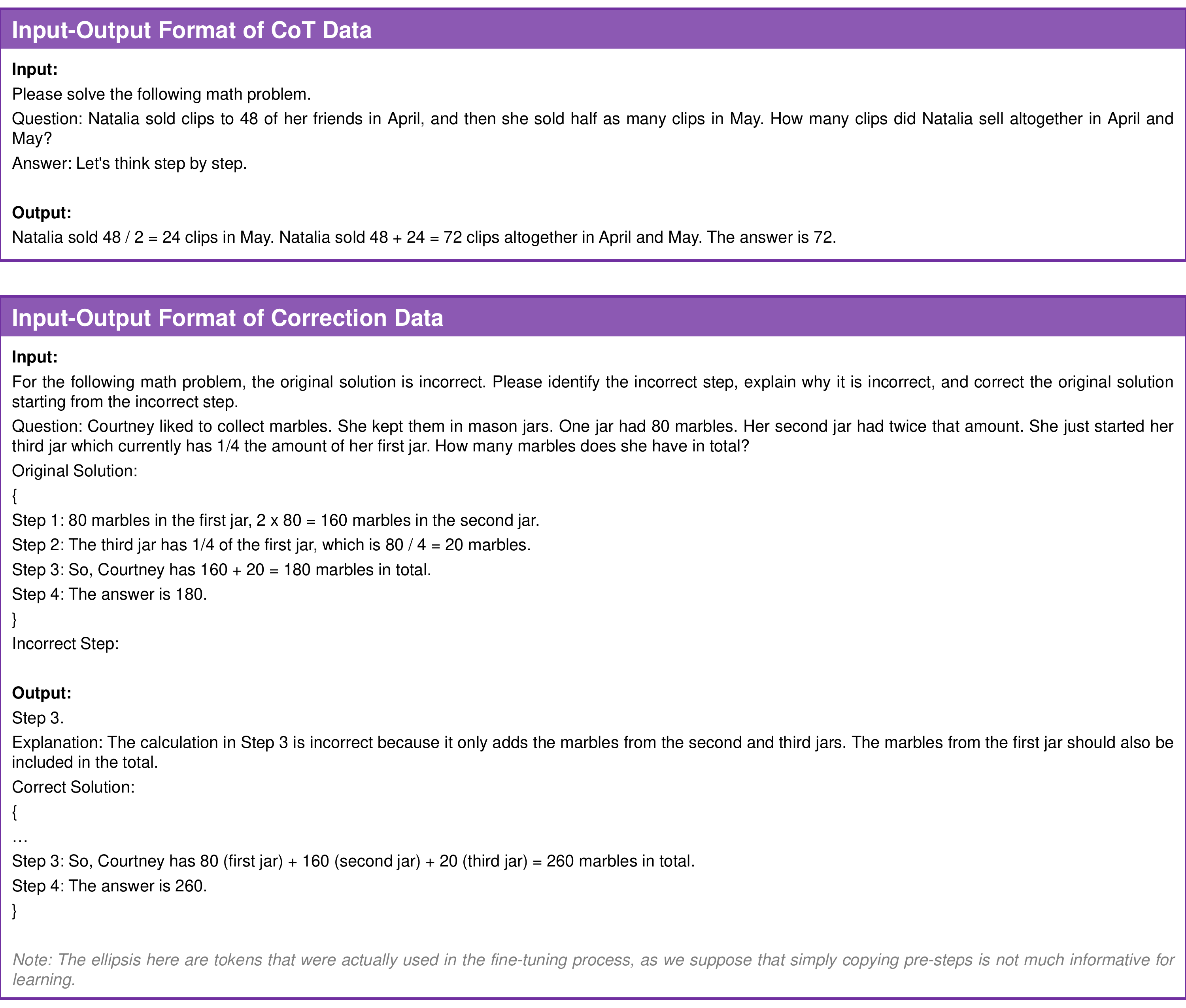}
    \caption{
    The input-output formats for our CoT data and correction data, respectively.
    The input part serves as a prompt and only the loss in the output part participates in the back-propagation.
    }
    \label{fig:input_output_format}
\end{figure*}

\subsection{Input-Output Formats for Fine-Tuning}\label{sec:ap_format}

Figure~\ref{fig:input_output_format} illustrate the input-output formats of CoT data and correction data, respectively.
Note that during the fine-tuning process, the input part serves as a prompt and only the loss in the output part participates in the back-propagation.

\subsection{Evolution Prompt}\label{sec:ap_evolve_prompt}

\begin{figure*}[t]
    \centering
    \includegraphics[width=.99\textwidth]{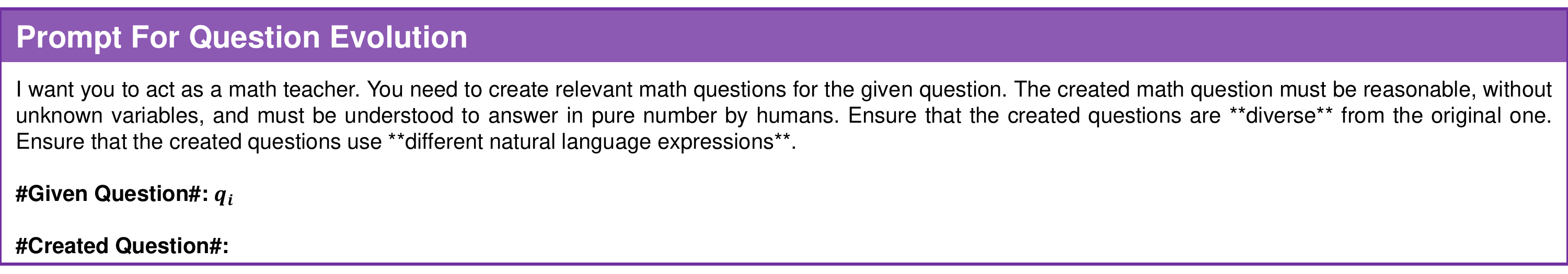}
    \caption{
    The prompt for evolving questions.
    }
    \label{fig:ap_evolve_prompt}
\end{figure*}

Figure~\ref{fig:ap_evolve_prompt} illustrates our prompt used for evolving new questions from the given seed question.

\section{More Details For Experimental Setup}

\subsection{Data Statistics}\label{sec:ap_data}

\begin{table}[t]
\renewcommand\arraystretch{1.2}
\Huge
\caption{Statistics of data sizes for the five tasks in our experiments (without question evolution).
}
\label{tab:dataset}
\centering
\resizebox{.7\linewidth}{!}{
\begin{tabular}{@{}cccc@{}}
\toprule
Task & CoT Data & Correction Data & Test Data \\ \midrule
GSM8K~\citep{cobbe2021training} & 32,421 & 12,523 & 1,319 \\
MATH~\citep{hendrycks2021measuring} & 20,009 & 6,306 & 5,000 \\
SVAMP~\citep{svamp2021} & - & - & 1,000 \\
ASDiv~\citep{asdiv2020} & - & - & 2,084 \\
CSQA~\citep{commonsenseqa2019} & 10,536 & 7,241 & 1,221 \\ \bottomrule
\end{tabular}
}
\end{table}

Table~\ref{tab:dataset} illustrates  basic statics about the tasks and data (without question evolution).

\subsection{Evaluation on ASDiv}\label{sec:ap_asdiv}

As mentioned in our setup, the original version of ASDiv contains 2,305 questions and part of them lead to non-numerical answers.
For instance, for the question ``Mrs. Hilt has two pennies, two dimes, and two nickels. Jacob has four pennies, one nickel, and one dime. Who has more money?'', the answer is the string value ``Mrs. Hilt'';
for the question ``Tessa has 4 apples. Anita gave her 5 more. She needs 10 apples to make a pie. Does she have enough to make a pie?'', the answer is a Boolean value ``False''.
As our models are trained on data derived from GSM8K where questions are all leading to numerical answers, it is reasonable that these models can not generate non-numerical answers.
Therefore, for evaluation on ASDiv, we filter out questions with non-numerical answers and finally leave 2,084 questions.
Specifically, for the question-answer pair in ASDiv, it will be filtered out if the answer can not be successfully recognized by the Python function $\mathrm{float(\cdot)}$.

\subsection{Data Construction For CSQA}\label{sec:ap_csqa}

\begin{figure*}[t]
    \centering
    \includegraphics[width=.99\textwidth]{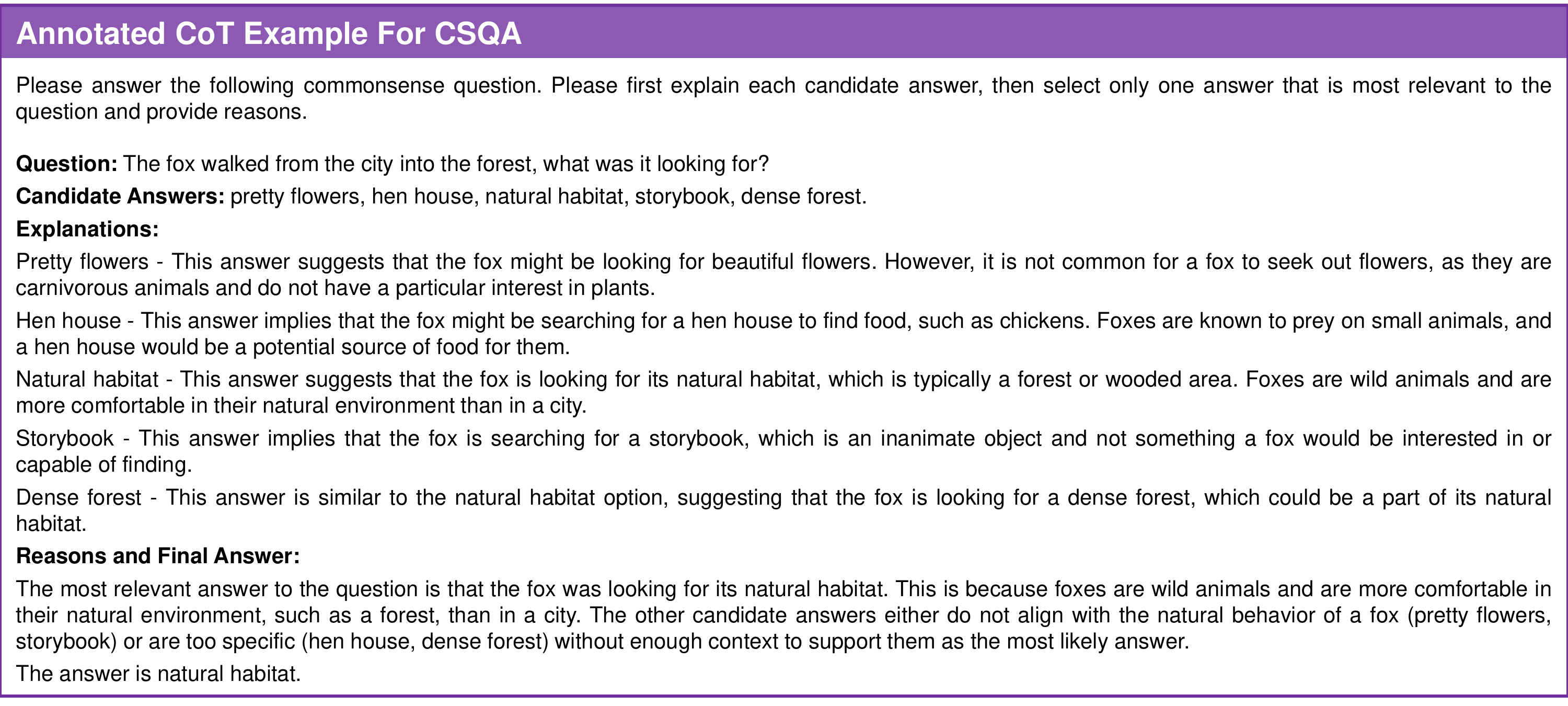}
    \caption{
    One annotated CoT example for CSQA.
    }
    \label{fig:csqa_example}
\end{figure*}

The original training examples in CSQA only contain the labeled final answers without rationales.
Therefore, we need to generate CoT for the training examples.
We first annotate rationales for four training examples.
Figure~\ref{fig:csqa_example} shows one annotated example.
Specifically, the CoT contain three parts: the explanation to each candidate answers, the predicted final answer, and the reason to choose this answer.
Then, we utilize GPT-4 to generate rationales for other training examples and filter out rationales that do not contain the correct final answers.
For generating correction data, we do not require GPT-4 to explicitly identify the position of mistake.
It is because the CoT for commonsense questions does not exhibit a clear step-wise manner, and our ablation study on math tasks have showed that this information is less influential to the final performance.

\subsection{Full Fine-Tuning Setting}\label{sec:ap_ft_set}
For fully fine-tuning LLaMA-2-70B and Llemma-34B, the learning rate is 1e-5 and the batch size is 128.
We fine-tune LLaMA-2-70B for 3 epochs and Llemma-34B for 2 epochs.
The evaluation results are reported on the final checkpoints.
Other setting are kept the same in Section~\ref{sec:ft_setting}.

\subsection{Another Round of Correction-Centric Evolution}\label{sec:ap_another_evolve}

To explore the scaling trend of \textsc{LeMa}, we take another round of correction-centric evolution to expand correction data.
The second round takes the same 10K seed questions as the first round.
The only difference is that we replace the vanilla model as the fine-tuned models from the first round to collect inaccurate reasoning paths.

\section{More Results and Analysis}\label{ap:more_results}

\subsection{Performances of Best Three Checkpoints}\label{sec:ap_best3}

\begin{table*}[t]
\renewcommand\arraystretch{1.2}
% \Huge
\caption{Performances of the \textbf{best three checkpoints} saved during the fine-tuning process and the average of three results.
}
\label{tab:best3}
\centering
\resizebox{.99\linewidth}{!}{
\begin{tabular}{@{}clcccc@{}}
\toprule
\multirow{2}{*}{Model} & \multicolumn{1}{c}{\multirow{2}{*}{Training}} & \multicolumn{2}{c}{GSM8K} & \multicolumn{2}{c}{MATH} \\ \cmidrule(l){3-6} 
 & \multicolumn{1}{c}{} & 1st / 2nd / 3rd & Avg. & 1st / 2nd / 3rd & Avg. \\ \midrule
\multirow{2}{*}{LLaMA-2-70B~\citep{touvron2023llama2}} & CoT Fine-Tuning & 81.4 / 81.3 / 81.1 & 81.3 & 23.6 / 23.2 / 23.2 & 23.2 \\
 & \cellcolor{gray!25} + Learning From   Mistakes & \cellcolor{gray!25}83.5 / 83.4 / 83.2 & \cellcolor{gray!25}83.4 (+2.1) & \cellcolor{gray!25}25.0 / 25.0 / 24.6 & \cellcolor{gray!25}24.9 (+1.7) \\ \midrule
\multirow{2}{*}{LLaMA-65B~\citep{touvron2023llama}} & CoT Fine-Tuning & 76.2 / 76.2 / 75.7 & 76.0 & 19.7 / 19.7 / 19.2 & 19.5 \\
 & \cellcolor{gray!25} + Learning From   Mistakes & \cellcolor{gray!25}77.9 / 77.3 / 77.2 & \cellcolor{gray!25}77.5 (+1.5) & \cellcolor{gray!25}20.8 / 20.3 / 20.2 & \cellcolor{gray!25}20.4 (+0.9) \\ \midrule
\multirow{2}{*}{CodeLLaMA-34B~\citep{rozière2023codellama}} & CoT Fine-Tuning & 68.8 / 68.5 / 68.2 & 68.5 & 19.1 / 19.0 / 18.9 & 19.0 \\
 & \cellcolor{gray!25} + Learning From   Mistakes & \cellcolor{gray!25}71.7 / 71.0 / 70.9 & \cellcolor{gray!25}71.2 (+2.7) & \cellcolor{gray!25}20.4 / 20.2 / 20.0 & \cellcolor{gray!25}20.2 (+1.2) \\ \midrule
\multirow{2}{*}{LLaMA-2-13B~\citep{touvron2023llama2}} & CoT Fine-Tuning & 62.9 / 62.7 / 62.7 & 62.8 & 12.2 / 11.9 / 11.8 & 12.0 \\
 & \cellcolor{gray!25} + Learning From   Mistakes & \cellcolor{gray!25}65.7 / 65.2 / 65.0 & \cellcolor{gray!25}65.3 (+2.5) & \cellcolor{gray!25}12.6 / 12.6 / 12.4 & \cellcolor{gray!25}12.5 (+0.5) \\ \midrule
\multirow{2}{*}{LLaMA-2-7B~\citep{touvron2023llama2}} & CoT Fine-Tuning & 52.6 / 52.5 / 52.5 & 52.5 & 8.7 / 8.5 / 8.5 & 8.6 \\
 & \cellcolor{gray!25} + Learning From   Mistakes & \cellcolor{gray!25}54.1 / 53.7 / 53.6 & \cellcolor{gray!25}53.8 (+1.3) & \cellcolor{gray!25}9.4 / 8.9 / 8.8 & \cellcolor{gray!25}9.0 (+0.4) \\ \bottomrule
\end{tabular}
}
\end{table*}

Table~\ref{tab:best3} shows the performances of the best three checkpoints saved during the fine-tuning process along with the average of three results.
It demonstrates that our main results are not caused by soem random disturbances during training.

\begin{figure*}[t]
     \centering
    \includegraphics[width=.85\textwidth]{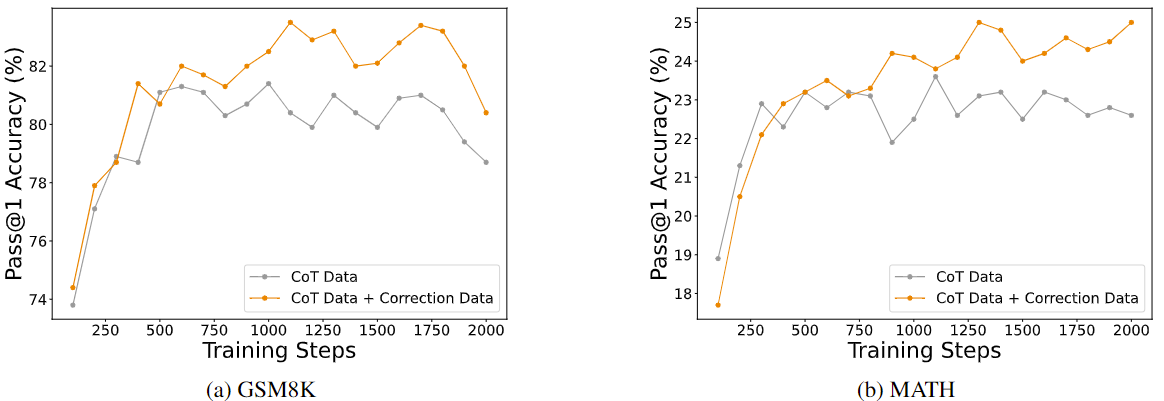}
     \caption{The performance curves of LLaMA-2-70B during 2,000 fine-tuning steps.
     }
     \label{fig:curve}
\end{figure*}

\subsection{Training Curves}\label{sec:ap_training_curve}

Figure~\ref{fig:curve} shows the performance curves of LLaMA-2-70B during 2,000 fine-tuning steps.
It shows that adding correction data leads to clear improvements during training.
These consistent improvements demonstrate that the effectiveness of our correction data is robust to the random disturbances during training.

\subsection{Comparison with SOTA Models}\label{sec:ap_sota}

\begin{table}[t]
\renewcommand\arraystretch{1.1}
% \Huge
\caption{Math reasoning performances of various LLMs.
}
\label{tab:lead}
\centering
\resizebox{.5\linewidth}{!}{
\begin{tabular}{@{}lcc@{}}
\toprule
\multicolumn{1}{c}{Model} & GSM8K & MATH \\ \midrule \midrule
\multicolumn{3}{c}{\textit{closed-source models}} \\
GPT-4~\citep{openai2023gpt4} & 92.0 & 42.5 \\
Claude-2~\citep{claude2} & 88.0 & - \\
Flan-PaLM-2~\citep{anil2023palm} & 84.7 & 33.2 \\
GPT-3.5-Turbo~\citep{gpt35turbo} & 80.8 & 34.1 \\
PaLM-2~\citep{anil2023palm} & 80.7 & 34.3 \\ \midrule \midrule
\multicolumn{3}{c}{\textit{open-source models}} \\
LLaMA-2-7B~\citep{touvron2023llama2} & 14.6 & 2.5 \\
Baichuan-2-7B~\citep{yang2023baichuan} & 24.5 & 5.6 \\
SQ-VAE-7B~\citep{wang2023guiding} & 40.0 & 7.0 \\
RFT-7B~\citep{yuan2023scaling} & 50.3 & - \\
Qwen-7B~\citep{qwen7b} & 51.6 & - \\
\cellcolor{gray!25}LLaMA-2-7B + \textsc{LeMa} (ours) & \cellcolor{gray!25}54.1 & \cellcolor{gray!25}9.4 \\
WizardMath-7B~\citep{luo2023wizardmath} & 54.9 & 10.7 \\
\cellcolor{gray!25}WizardMath-7B + \textsc{LeMa} (ours) & \cellcolor{gray!25}55.9 & \cellcolor{gray!25}11.9 \\
LLaMA-2-13B~\citep{touvron2023llama2} & 28.7 & 3.9 \\
SQ-VAE-13B~\citep{wang2023guiding} & 50.6 & 8.5 \\
Baichuan-2-13B~\citep{yang2023baichuan} & 52.8 & 10.1 \\
RFT-13B~\citep{yuan2023scaling} & 54.8 & - \\
WizardMath-13B~\citep{luo2023wizardmath} & 63.9 & 14.0 \\
\cellcolor{gray!25}LLaMA-2-13B + \textsc{LeMa} (ours) & \cellcolor{gray!25}65.7 & \cellcolor{gray!25}12.6 \\
MetaMath-13B~\citep{yu2023metamath} & 72.3 & 22.4 \\
\cellcolor{gray!25}MetaMath-13B + \textsc{LeMa} (ours) & \cellcolor{gray!25}73.2 & \cellcolor{gray!25}22.7 \\
LLaMA-2-70B~\citep{touvron2023llama2} & 56.8 & 13.5 \\
RFT-70B~\citep{yuan2023scaling} & 64.8 & - \\
WizardMath-70B~\citep{luo2023wizardmath} & 81.6 & 22.7 \\
MuggleMath-70B~\citep{li2023query} & 82.3 & - \\
MetaMath-70B~\citep{yu2023metamath} & 82.3 & 26.6 \\
\cellcolor{gray!25}LLaMA-2-70B + \textsc{LeMa} (ours) & \cellcolor{gray!25}83.5 & \cellcolor{gray!25}25.0 \\
\cellcolor{gray!25}WizardMath-70B + \textsc{LeMa} (ours) & \cellcolor{gray!25}84.2 & \cellcolor{gray!25}\textbf{27.1} \\ 
\cellcolor{gray!25}MetaMath-70B + \textsc{LeMa} (ours) & \cellcolor{gray!25}\textbf{85.4} & \cellcolor{gray!25}26.9 \\ \bottomrule
\end{tabular}
}
\end{table}

Table~\ref{tab:lead} contains the comparison with more SOTA models.
Another interesting finding in Table~\ref{tab:lead} is that the performance of LLaMA-2-70B + \textsc{LeMa} can be comparable with MuggleMath-70B~\citep{li2023query} and MetaMath-70B~\citep{yu2023metamath}.
Note that these two specialized LLMs also take the LLaMA-2-70B as the backbone model while their training data sizes are much larger than \textsc{LeMa}:
MuggleMath has $\sim$220K CoT data and MetaMath has $\sim$400K CoT data, while \textsc{LeMa} only has $\sim$70K CoT + correction data for math problems.
This comparison further supports the non-homogeneous effectiveness between CoT data and correction data.

\begin{figure*}[t]
    \centering
    \includegraphics[width=.95\textwidth]{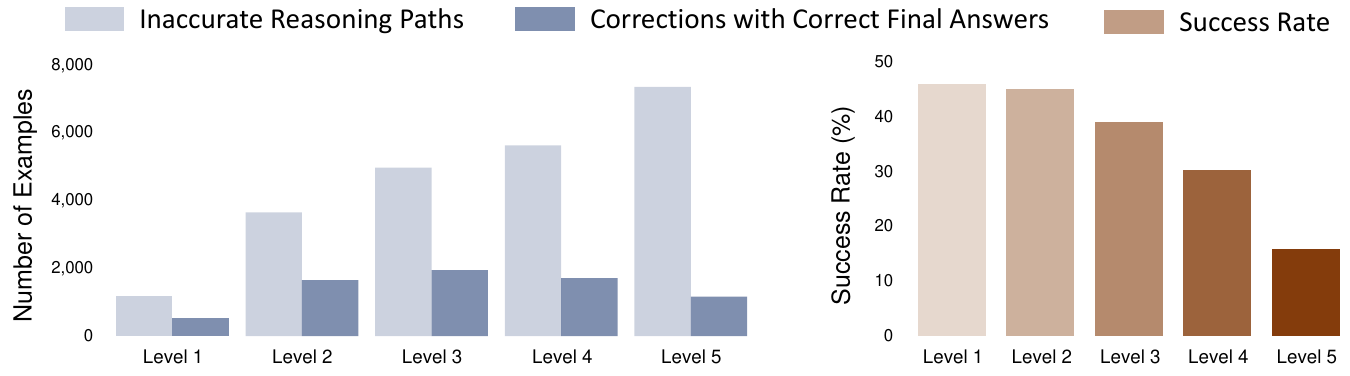}
    \caption{
    Statistics of generated correction data according to different difficulty levels in MATH.
    \textbf{Left:} The number of collected inaccurate reasoning paths and generated corrections with correct final answers under different difficulty levels.
    \textbf{Right:} The success rate for correcting inaccurate reasoning paths under different difficulty levels.
    }
    \label{fig:level_stat}
\end{figure*}

\begin{figure*}[t]
    \centering
    \includegraphics[width=.5\textwidth]{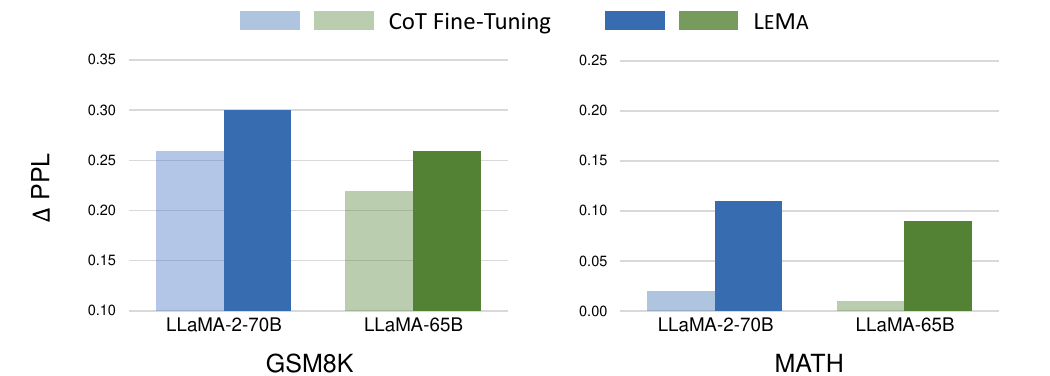}
    \caption{
    The differences between the PPLs ($\Delta\mathrm{PPL}$) on mistaken CoT and correct CoT.
    A higher difference indicate that the model can better avoid the mistakes.
    }
    \label{fig:ppl_diff}
\end{figure*}

\subsection{Ablations of Correction Information}

\begin{figure}[t]
    \centering
    \includegraphics[width=.49\textwidth]{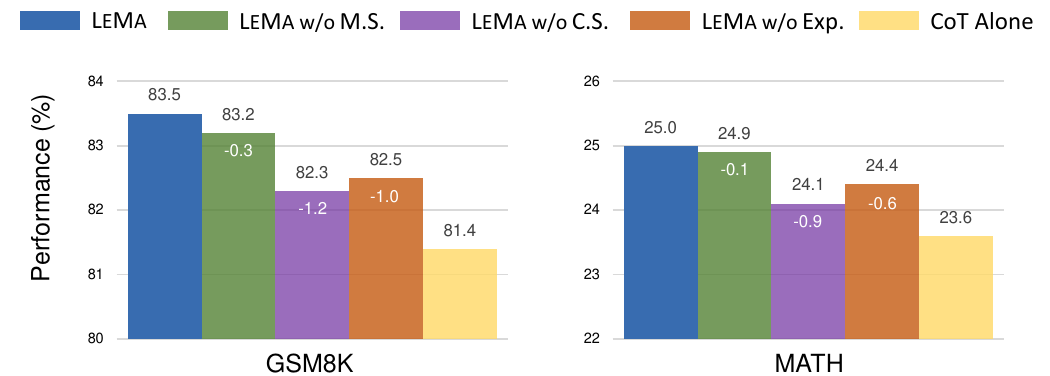}
    \caption{
    Performance of \textsc{LeMa} with ablations on correction information.
    The backbone LLM is LLaMA-2-70B.
    For each ablation setting, we mark the influence on performance compared to the default setting of \textsc{LeMa}.
    }
    \label{fig:ablation}
\end{figure}

\paragraph{The explanations and corrected reasoning paths play important roles in \textsc{LeMa}.}
As introduced in Section~\ref{sec:method_data}, our correction data mainly contains three pieces of information: the mistake step (M.S.), the corrected solution (C.S.), and the explanation to the mistake (Exp.).
To evaluate their individual contribution to the \textsc{LeMa} performance, we separately omit each information in our correction data.
Figure~\ref{fig:ablation} shows the results:
the performance of \textsc{LeMa} drops significantly without the corrected solution or the explanation, while omitting the mistake step shows less influence to the performance.
We suppose it is because the corrected solution and the explanation have implicitly informed which step is incorrect.
Therefore, it could be less influential to make the model explicitly identify the position of mistake.

\subsection{Additional Analysis to \textsc{LeMa}}\label{sec:ap_more_analysis}

\paragraph{\textsc{LeMa} can still bring improvements to CoT fine-tuning if the distributions of questions are controlled the same.}
In our default setting, correction data contains more challenging questions that can not be easily solved by various LLMs.
This leads to a distribution shift on the difficulty of questions in training data.
As \citet{wang2023lets} indicated that this distribution shift can also benefit fine-tuning LLMs, we also mitigate the influence from question distribution shift to further clarify the effectiveness of \textsc{LeMa}.
Our ablation setting CoT-45K can be used to clarify this point: its additional CoT data are just converted from correction data, thus the question distributions of CoT-45K and our default \textsc{LeMa}-45K are exactly the same.
Therefore, the results in Figure~\ref{fig:datasize} under 45K data size demonstrate that \textsc{LeMa} still outperforms CoT-alone fine-tuning when the influence from question distribution shift is kept the same.

\paragraph{QLoRA fine-tuning cannot fully ``digest'' a large amount of correction data.}
As shown in Figure~\ref{fig:evolve_trend}, as the correction data expands, the gap between full-fine-tuning and QLoRA fine-tuning increases.
Such an observation is not well aligned with the conclusions of some existing work.
Some work indicated that if the model size is large enough, parameter-efficient fine-tuning (PEFT) can achieve comparable performance with fine-tuning ~\citep{lester2021power, an2022inputtuning, sun2023comparative, su2023exploring, loravsfinetune2023}.
We suppose the property of correction data causes the inconsistency in observations.
Specifically, correction data is just auxiliary data that do not directly contribute to the in-task training.
We suppose that models with PEFT can ``eat'' a large amount of correction data but cannot fully ``digest'' them.
As a results, the training on correction data with PEFT might not effectively contribute to the forward reasoning process.
% not effectively ``learn'' from them and transfer to the forward reasoning process.

% \paragraph{Larger LLMs can better benefit from \textsc{LeMa}.}
% For applying \textsc{LeMa} on specialized LLMs, we find that larger specialized models obtain more gains than smaller ones.
% As shown in Table~\ref{tab:lead}, MetaMath-70B improves 3.1\% pass@1 accuracy score on GSM8K while MetaMath-13B only has 0.9\% accuracy improvement;
% WizardMath-70B improves 4.4\% on MATH while WizardMath-7B only improves 1.2\%.
% Along with the performance of LLaMA-2-7B with controlled data size for \textsc{LeMa} (Figure~\ref{fig:datasize}), these comparisons suggest that larger LLMs are better at learning from mistakes.

\paragraph{The comparison learned in the correction data also influences the CoT generation.}
During training on the correction data, LLMs could be aware of the comparison between the correct and incorrect CoT.
We suppose such kind of comparison can take effect during CoT generation.
% \paragraph{\textsc{LeMa} makes LLMs better avoid mistakes that have been taught.}
% If the model has learned to avoid certain mistakes, it should have a higher perplexity (PPL) on the mistaken CoT than on the correct CoT.
Based on this intuition, we evaluate the differences between PPLs defined as follows,
\begin{equation}
    \Delta\mathrm{PPL}(\mathcal{C};\theta)=\\\nonumber\frac{1}{||\mathcal{C}||}\sum_{(q_{i}, \widetilde{r_{i}}, c_{i}) \in \mathcal{C}} [\mathrm{PPL}(\widetilde{r_{i}}|q_{i};\theta) -  \mathrm{PPL}(r_{i}|q_{i};\theta)],
\end{equation}
where $\mathcal{C}$ is a set of correction data, $\theta$ represents the model parameters after fine-tuning, $\mathrm{PPL}(y|x;\theta)$ returns the perplexity on $y$ with $x$ as the context, $\widetilde{r_{i}}$ is one mistaken CoT for the question $q_{i}$, and $r_{i}$ is the correct CoT extracted from the correction $c_{i}$.
We calculate $\Delta\mathrm{PPL}$ for fine-tuned LLaMA-2-70B and LLaMA-65B, based on the correction data for GSM8K and MATH.
Figure~\ref{fig:ppl_diff} shows $\Delta\mathrm{PPL}$ for different fine-tuned models.
It shows that \textsc{LeMa} consistently leads to a higher $\Delta\mathrm{PPL}$ than CoT-alone fine-tuning.
% This comparison reflects that after the \textsc{LeMa} training process, these fine-tuned LLMs have indeed learned to avoid mistakes present in correction data.
% This comparison reflects that models trained by \textsc{LeMa} has indeed learned to avoid mistakes present in correction data.

\subsection{Further Analysis on Corrector}\label{sec:ana_corrector}

In our default setting, we take GPT-4 as the corrector model and our human evaluation in Section~\ref{sec:method_data} supports this choice.
In the following, we provide further analysis on the choice and behavior of the corrector model.
Specifically, we want to answer the following research questions:
\textbf{RQ1:} Can we use a less powerful model as the corrector model?
\textbf{RQ2:} How well does GPT-4 perform in self-correction?
\textbf{RQ3:} How well does GPT-4 correct inaccurate reasoning paths for challenging questions?

\paragraph{Less powerful models are not suitable for generating corrections.}
Despite GPT-4, we have also tried leveraging GPT-3.5-Turbo as the corrector model and assess the quality of generated corrections.
% Our human evaluation found that nearly half of corrections generated by GPT-3.5-Turbo are of poor quality.
We take another round of human evaluation on 20 corrections generated by GPT-3.5-Turbo and find that nearly half are of poor quality.
Therefore, we just call GPT-4 for correction generation although it is much more expensive than GPT-3.5-Turbo.
We believe it is a valuable research direction to explore how to generate high-quality corrections without GPT-4.

\paragraph{GPT-4 can correct its own mistakes but with a low success rate.}
Specifically, for 2,696 inaccurate reasoning paths generated by GPT-4 on MATH training set, we finally get 217 corrections with correct final answers.
It means that GPT-4 only achieves 8.0\% success rate for self-correction.
Compared with this low success rate for self-correction, GPT-4 can more effectively correct mistakes from less powerful models, such as LLaMA-2-70B (37.5\% success rate on MATH) and GPT-3.5-Turbo (26.9\% success rate on MATH).
One possible reason for the low success rate of self-correction is that the mistakes generated by GPT-4 are from more challenging questions, thus these mistakes are naturally harder for correcting.

\paragraph{GPT-4 still struggles to correct inaccurate reasoning paths for challenging questions.}
The math problems in MATH can be categorized into five levels of difficulty: Level 1 for the easiest problems and Level 5 for the most challenging ones.
Figure~\ref{fig:level_stat} shows statistics of our correction data on MATH according to different difficulty levels.
As the difficulty increased from Level 1 to Level 5, the number of collected inaccurate reasoning paths increased, while the number of correct corrections (i.e., corrections for which the final answer is correct) first increases and then decreases.
We also calculate the success rate for correcting mistakes under each difficulty level, dividing the number of correct corrections by the total number of collected reasoning paths.
Figure~\ref{fig:level_stat} shows that the success rate significantly drops with increasing the difficulty.
These statistics reveals that there is still huge room for improving contemporary LLMs on correcting mistakes.

\end{document}